\documentclass[journal, a4paper, twocolumn]{IEEEtran}
\usepackage{ifpdf}

\ifCLASSINFOpdf
   \usepackage[pdftex]{graphicx}
 \else
   \usepackage[dvips]{graphicx}
 \fi

\usepackage{booktabs}
\usepackage{paralist} 
\usepackage{mathrsfs,amsfonts,amsmath,amsthm}
\usepackage{multicol}
\usepackage{subfigure}
\usepackage{graphicx}
\usepackage{multirow}
\usepackage{subfigure}
\usepackage{algorithm}
\usepackage{algorithmic}
\usepackage{setspace}
\usepackage{amssymb}

\newtheorem{theorem}{Theorem}
\newtheorem{lemma}{Lemma}
\newtheorem{corollary}{Corollary}

\theoremstyle{definition}

\theoremstyle{definition}

\theoremstyle{definition}
\newtheorem{remark}{Remark}

\theoremstyle{definition}

\usepackage{siunitx}
\usepackage{color}

\def \x {\mathbf{x}}

\def \w {\mathbf{w}}

\def \b {\mathbf{b}}

\def \g {\mathbf{g}}
\def \e {\mathbf{e}}
\def \s {\mathbf{s}}
\def \u {\mathbf{u}}
\def \v {\mathbf{v}}
\def \Ee {\mathbf{E}}

\def \X {\mathbf{X}}

\def \G {\mathbf{G}}
\def \F {\mathbf{F}}

\def \V {\mathbf{V}}
\def \P {\mathbf{P}}
\def \S {\mathbf{S}}
\def \U {\mathbf{U}}
\def \I {\mathbf{I}}

\def \A {\mathbf{A}}
\def \C {\mathbf{C}}
\def \D {\mathbf{D}}
\def \Y {\mathbf{Y}}

\def \T {\mathbf{T}}

\def \R {\mathbb{R}}

\def \E {\mathbb{E}}
\def \Pr {\mathbb{P}}
\def \Tr {\mathrm{Tr}}
\def \diag {\mathbb{D}}
\def \bS {\mathbf{\Sigma}}

\def \l {\ell}

\usepackage{url}
\usepackage[numbers]{natbib}

\begin{document}

%
\title{Effective Data-aware Covariance Estimator from Compressed Data}
%
%
%

\author{Xixian Chen, Haiqin Yang,~\IEEEmembership{Member,~IEEE,} Shenglin Zhao,~\IEEEmembership{Member,~IEEE,}
Michael~R.~Lyu,~\IEEEmembership{Fellow,~IEEE,}
and Irwin~King~\IEEEmembership{Fellow,~IEEE}
\thanks{
X.~Chen and S.~Zhao are with the Youtu Lab, Tencent,
Shenzhen, China, postal code: 518057.  \{xixianchen, henryslzhao\}@tencent.com}
\thanks{
M.~R.~Lyu, and I.~King are with Shenzhen Research Institute, The Chinese University of Hong Kong, Shenzhen, China, postal code: 518057, and Department of Computer Science and  Engineering, The Chinese University of Hong Kong, Shatin, N.T., Hong Kong.  \{lyu, king\}@cse.cuhk.edu.hk}
\thanks{H.~Yang is the corresponding author and is affiliated with Meitu (China), Hong Kong and Department of Computing, Hang Seng University of Hong Kong.  haiqin.yang@gmail.com.}
\thanks{The work was fully supported by the Research Grants Council of the Hong Kong Special Administrative Region, China (No. CUHK 14208815, No. CUHK 14234416, and Project No.~UGC/IDS14/16).
}
\thanks{Manuscript received xx xx, 201x; revised xx xx, 201x.}
}

%
%

\markboth{IEEE Transactions on Neural Networks and Learning Systems,~Vol.~, No.~, ~201X}%
{DACE}
%



\maketitle
\allowdisplaybreaks
\begin{abstract}
Estimating covariance matrix from massive high-dimensional and distributed data is significant for various real-world applications.  In this paper, we propose a data-aware weighted sampling based covariance matrix estimator, namely DACE, which can provide an unbiased covariance matrix estimation and attain more accurate estimation under the same compression ratio.  Moreover, we extend our proposed DACE to tackle multiclass classification problems with theoretical justification and conduct extensive experiments on both synthetic and real-world datasets to demonstrate the superior performance of our DACE.
\end{abstract}


\IEEEpeerreviewmaketitle
\section{Introduction}\label{sec:FROH_intro}
\IEEEPARstart{C}{ovariance} matrix, absorbing the second order information of data {points}, plays a significant role in many machine learning and statistics applications~\cite{feller1966introduction}.  For example, {the estimated covariance matrix plays the role of dimension reduction or denoising for Principal Component Analysis (PCA)~\cite{zou2006sparse}, Linear Discriminant Analysis (LDA) and Quadratic Discriminant Analysis (QDA)~\cite{anzai2012pattern}, etc.  Via an estimated noise covariance matrix, Generalized Least Squares (GLS) regression can attain the best linear estimator~\cite{kariya2004generalized}.}  The Independent Component Analysis (ICA) relies on the covariance matrix for pre-whitening~\cite{hyvarinen2004independent}.  The Generalized Method of Moments (GMM)~\cite{hansen1982large} improves the effectiveness of the model by estimating a precise covariance matrix.  Many real-world applications, such as gene relevance networks~\cite{butte2000discovering,schafer2005empirical}, modern wireless communications~\cite{tulino2004random}, array signal processing~\cite{abrahamsson2007enhanced}, and policy learning~\cite{DBLP:journals/ftrob/DeisenrothNP13}, also rely on directly estimating the covariance matrix~\cite{bartz2016advances}. 

Nowadays, large and high-dimensional data are routinely generated in various distributed applications, such as sensor networks, surveillance systems, and distributed  databases~\cite{ha2015robust,haupt2008compressed,shi2014correlated}.  The communication cost becomes challenging because the distributed data needs to be transmitted to a fusion center from remote sites, requiring tremendous bandwidth and power consumption~\cite{abbasi2016toward,srisooksai2012practical}.   One effective solution is to utilize the compressed data, i.e., {projecting the original data to a small-size one via a Gaussian matrix}, where the space cost, the computational cost, and the communication cost can be reduced significantly to linearly depend on the projected size.  However, the above solution suffers from two critical drawbacks.  First, projecting to a Gaussian matrix is inefficient compared with computing sparse projection matrices~\cite{li2006very}, structured matrices~\cite{chen2017frosh}, or sampling matrix~\cite{drineas2006subspace}.  Second, applying the same projection matrix to all data points cannot recover the original covariance matrix precisely.  Current theoretical investigation and empirical results show that {even the size of the samples with a fixed dimension increases to infinity, the estimator cannot recover the target covariance matrix}~\cite{pourkamali2015preconditioned,anaraki2014memory,azizyan2015extreme, gleichman2011blind}. 

To tackle the above challenges, we propose a Data-Aware Convariance matrix Estimator, namely DACE, to leverage different projection matrices for {each data point}.  It is known that without statistical assumptions or low-rank/sparsity structural assumptions on the distribution of the data, our DACE can achieve consistent covariance matrix estimation.  By a crafty designed weighted sampling scheme, we can compress the data and recover the covariance matrix in the center efficiently and precisely.  We summarize our contributions as follows:
\begin{itemize}
\item First, we propose a data-aware covariance matrix estimator by a weighted sampling scheme.  This is different from existing data-oblivious projection methods~\cite{pourkamali2015estimation,pourkamali2015preconditioned,anaraki2014memory,azizyan2015extreme}.  By exploiting the most important entries, our strategy requires considerably fewer entries to achieve the same estimation precision. 

\item Second, we rigorously prove that our DACE is an unbiased covariance estimator.  Moreover, our DACE can {achieve more accurate estimation precision and consume less time cost than existing methods under the same compression ratio.}  The theoretical justification is verified in both synthetic and real-world datasets.


\item Third, we extend our DACE to tackle the multiclass classification problem and provide both {theoretical justification} and empirical evaluation.  The compact theoretical result and the superior empirical performance imply that the covariance matrix estimated from compressed data indeed guarantees the intrinsic properties of data and can be applied in {various} down-stream applications. 

\end{itemize} 


\section{Problem Definition and Related Work} \label{sec:related}

\subsection{Notations and Problem Definition} 
Following the notations defined in~\cite{chen2017toward}, {given $n$ distributed data in $g$ remote sites}, $\X=[\x_1, \ldots, \x_n]$, where $\x_i\in\R^d$ and $g\ll n$, the corresponding covariance matrix can be computed by $\C=\frac{1}{n}\X\X^T-\bar{\x}\bar{\x}^T$, where  $\bar{\x}=\frac{1}{n}\sum_{i=1}^n\x_i$ can be exactly computed in the fusion center through $\bar{\x}=\frac{1}{n}\sum_{j=1}^g\g_j$, where  $\g_j\in\R^d$ represents the summation of all data points in the $j$-th remote site before being compressed.  {Hence, without loss of generality, we can assume zero-mean, i.e., $\bar{\x}=\mathbf{0}$.} 

Now, we define the procedure of covariance matrix recovery as follows: given data $\X$ and specific designed sampling matrices, $\{\S_i\}_{i=1}^n\in \R^{d\times m}$, where $m\ll \{d, n\}$, the original data is compressed via $\S_i^T\x_i$ and transmitted to the fusion center while the covariance matrix of the original data is recovered by a transformation only via $\S_i$.  The question is how to design the sampling matrices to guarantee the estimated covariance matrix as precisely as possible.  

\subsection{Related Work}
Various randomized algorithms have been proposed to tackle the above problem and can be divided into three main streams:
\begin{compactitem}
\item {\em Independent Projection.} The \textit{Gaussian-inverse} method~\cite{qi2012invariance} applies a Gaussian matrix $\S_i$ to compress each data point and recovers the data via $\S_i(\S_i^T\S_i)^{-1}(\S_i^T\x_i)$.  The information of all entries in each data vector is very likely to be acquired uniformly and substantively because $\S_i(\S_i^T\S_i)^{-1}\S_i^T$ is an $m$-dimensional orthogonal projection, whose projection spaces are uniformly and randomly drawn.  Hence, $\frac{1}{n}\sum_{i=1}^n\S_i(\S_i^T\S_i)^{-1}\S_i^T\x_i\x_i^T\S_i(\S_i^T\S_i)^{-1}\S_i^T$ is expected to constitute an accurate and consistent covariance matrix estimation up to a known scaling factor~\cite{azizyan2015extreme}.  However, computing the Gaussian matrix is computational burden because the Gaussian matrix is dense and unstructured.  Moreover, the matrix inverse operation requires much computational time and memory cost.  A biased estimator $\frac{1}{n}\sum_{i=1}^n\S_i\S_i^T\x_i\x_i^T\S_i\S_i^T$ is then presented in~\cite{anaraki2014memory} by applying a sparse matrix $\S_i$ to avoid computing the matrix inversion.  This strategy is less accurate because $\S_i\S_i^T$ approximates only an $m$-dimensional random orthogonal projection.  Moreover, its performance is guaranteed only on data that satisfy a certain statistical assumption, e.g., Gaussian distribution.  An unbiased estimator~\cite{pourkamali2015estimation} is then proposed to adopt an unstructured sparse matrix to construct the projection.  The method is computational inefficient and fails to afford error bounds to trade off the estimation error and the compression ratio.  To improve computational efficiency, the strategy of sampling \textit{without replacement} has been employed to $\S_i$.  However, this method recovers the data via $\S_i^T\x_i$, which is poor because $\S_i\S_i^T$ is an $m$-dimensional orthogonal projection drawn only from $d$ deterministic orthogonal spaces/coordinates and removes $(d-m)$ entries of each vector.  To retain the accuracy, the Hadamard matrix~\cite{tropp2011improved} is applied in~\cite{pourkamali2015preconditioned} before sampling, which flattens out all entries, particularly those with large magnitudes, to all coordinates.  Though the proposed uniform sampling scheme can capture sufficient information embedded in all entries, it fails to capture the information uniformly in all coordinates of each vector because the Hadamard matrix involves deterministic orthogonal projection.  Hence, it requires numerous samples to obtain sufficient accuracy~\cite{pourkamali2015preconditioned}.  Overall, existing independent projection methods cannot capture the most valuable information sufficiently.   
\item {\em Projection via a low-rank matrix.}  A representative work~\cite{halko2011finding, chen2015training} is to improve the approximation precision by projecting the original data via a low-dimensional data-aware matrix $\X\widehat{\S}$, where $\widehat{\S}$ is a random projection matrix and $\X$ must be a low-rank matrix.  This method has to take one extra pass through all entries in $\X$ to compute $\X\widehat{\S}$.  Theoretical and empirical investigation shows that a single projection matrix for all data points cannot consistently and accurately estimate the covariance matrix~\cite{azizyan2015extreme}.  The problem of inconsistent covariance estimation and the restriction of low-rank matrix assumption also exist  in~\cite{mroueh2016co,wu2016single} for fast approximating matrix products in a single pass.
\item {\em Sampling in a whole.}  Other existing  methods~\cite{drineas2006fast,holodnak2015randomized,papailiopoulos2014provable,woodruff2014sketching} leverage column-based sampling to apply the column norms or leverage scores in the sampling probabilities matrix, while in~\cite{achlioptas2013near,achlioptas2007fast,woodruff2014sketching}, element-wise sampling is applied in the entire matrix.  These methods adopt various sampling distributions to sample entries from a matrix.  However, they require one or more extra passes over data because computing the sampling distributions requires to observe all data.  Moreover, the sampling probabilities are created for matrix approximation and cannot be trivially extended to covariance matrix estimation because it is not allowed to obtain the exact covariance matrix in advance.  Note that although the uniform sampling is a simple one-pass algorithm for matrix approximation, the structural non-uniformity in the data makes it perform poorly~\cite{pourkamali2015preconditioned}.
\end{compactitem}
Other than randomized algorithms, researchers also establish theory to recover the covariance matrix from given data~\cite{bioucas2014covalsa,cai2015rop,chen2013exact,dasarathy2015sketching}.  However, these methods are only applicable when the covariance matrix is low-rank, sparse, or follows a certain statistical assumption, {and restrict their application potentials.}

\begin{algorithm}[htbp]  
\setlength{\abovecaptionskip}{-0.13cm}
\setlength{\belowcaptionskip}{-0.90cm}
\setlength{\parskip}{-2cm}
\caption{Data-aware Covariance Estimator (DACE)}
\label{alg:mine}   
\begin{algorithmic}[1] 
\REQUIRE ~~\\ 
Data $\X\in\R^{d\times n}$, sampling size $m$, and $0 < \alpha < 1$.  
\ENSURE ~~\\ 
Estimated covariance matrix $\C_e\in\R^{d\times d}$. 
\STATE   Initialize $\Y\in\R^{m\times n}$, $\T\in\R^{m\times n}$, $\v\in \R^{n}$, and $\w\in \R^{n}$ with $\mathbf{0}$.
\FOR{all $i \in [n]$}
\STATE Load $\x_i$ into  memory, let $v_i=\|\x_i\|_1=\sum_{k=1}^d|x_{ki}|$ and $w_i=\|\x_i\|_2^2=\sum_{k=1}^dx_{ki}^2$
\FOR {all $j \in [m]$} 
\STATE  Pick $t_{ji}\in [d]$ with $p_{ki}\equiv\Pr(t_{ji}=k)=\alpha\frac{|x_{ki}|}{v_i}+(1-\alpha)\frac{x_{ki}^2}{w_i}$, and let $y_{ji}=x_{t_{ji}i}$ 
\ENDFOR
\ENDFOR
\STATE Pass the compressed data $\Y$, sampling indices $\T$, $\v$, $\w$, and $\alpha$
to the fusion center. 
\FOR{all $i \in [n]$} 
\STATE Initialize $\S_i\in \R^{d\times m}$ and $\P\in\R^{d\times n}$ with $\mathbf{0}$
\FOR {all $j \in [m]$}
\STATE Let $p_{t_{ji}i}=\alpha\frac{|y_{ji}|}{v_i}+(1-\alpha)\frac{y_{ji}^2}{w_i}$, and  $s_{t_{ji}j,i}=\frac{1}{\sqrt{mp_{t_{ji}i}}}$
\ENDFOR
\ENDFOR
\STATE Compute $\C_e$ as defined in Eq.~(\ref{eq:theorem21}).  
\end{algorithmic}  
\end{algorithm}

\section{Our Proposal} \label{sec:approach}

\subsection{Method and Algorithm}
Our proposed DACE utilizes data-aware weighted sampling matrices $\{\S_i\}_{i=1}^n$ to compress each data via $\S_i^T\x_i$ and then back-project the compressed data into the original space via $\S_i\S_i^T\x_i$.  The estimated covariance matrix is computed by 
\begin{eqnarray}
&\qquad\C_{e}=\widehat{\C}_1-\widehat{\C}_2 \label{eq:theorem21},\quad  \mbox{where~~}b_{ki}=\frac{1}{1+(m-1)p_{ki}}, \\ 
&\widehat{\C}_1=\frac{m}{nm-n}\sum_{i=1}^{n}\S_i\S_i^T\x_i\x_i^T\S_i\S_i^T,\label{eq:Chat_1}\\\label{eq:Chat_2}
&\widehat{\C}_2=\frac{m}{nm-n}\sum_{i=1}^{n}\diag(\S_i\S_i^T\x_i\x_i^T\S_i\S_i^T)\diag(\b_i)
\end{eqnarray}
In Eq.~(\ref{eq:theorem21}), at most $m$ entries in each $\b_i$ have to be calculated because each $\S_i\S_i^T\x_i\x_i^T\S_i\S_i^T$ contains at most $m$ non-zero entries in its diagonal. 

Algorithm~\ref{alg:mine} outlines the flow of DACE: Steps 1 to 7 show the procedure of compressing the data $\X$ to $\Y$,  where each entry is retained according to the probability proportional to the combination of its relative absolute value and the square value. {Step 8 describes the communication procedure to transmit the compressed data from all the remote sites to the fusion center}.  Steps 9 to 14 reveal the construction of an unbiased covariance matrix estimator in the fusion center from the compressed data.  {It is shown that only one pass is required to load all data from the external space into the memory, which reveals the applicability of our DACE for streaming data.}


\subsection{Primary Provable Results}
The following theorem shows that our proposed DACE can attain an unbiased estimator for the target covariance matrix.  
\begin{theorem} \label{the:unbiased}
Assume $\X\in \R^{d\times n}$ and the sampling size $2\leq m<d$.  $m$ entries are sampled from each $\x_i\in\R^d$ with replacement by running Algorithm~\ref{alg:mine}. Let $\{p_{ki}\}_{k=1}^d$ and $\S_i\in\R^{d\times m}$ denote the sampling probabilities and sampling matrix, respectively. Then, the unbiased estimator for the target covariance matrix $\C=\frac{1}{n}\sum_{i=1}^{n}\x_i\x_i^T=\frac{1}{n}\X\X^T$ can be recovered by Eq.~(\ref{eq:theorem21}).
\end{theorem}
The detailed proof is provided in Appendix~\ref{app:proof_unbiased}.  The estimation error can be also bounded by the following theorem:
\begin{theorem} \label{the:error}
{Let $\X$, $m$, $\C$ and $\C_e$ be defined as in Theorem~\ref{the:unbiased}.} If the sampling probabilities satisfy $p_{ki}=\alpha\frac{|x_{ki}|}{\|\x_i\|_1}+(1-\alpha)\frac{x_{ki}^2}{\|\x_i\|_2^2}$ with $0<\alpha<1$ for all $k\in [d]$ and $i\in [n]$, then with probability at least $1-\eta-\delta$, 
\begin{align}
\|\C_e-\C\|_2\leq \log(\frac{2d}{\delta})\frac{2R}{3}+\sqrt{2\sigma^2\log(\frac{2d}{\delta})} \label{eq:theorem22},
\end{align}
where  $R=\max_{i\in [n]}\left[\frac{7\|\x_i\|_2^2}{n}+\log^2(\frac{2nd}{\eta})\frac{14\|\x_i\|_1^2}{nm\alpha^2}\right]$, and
$\sigma^2=\sum_{i=1}^n\! \left[\frac{8\|\x_i\|_2^4}{n^2m^2(1-\alpha)^2}\!+\!\frac{4\|\x_i\|_1^2\|\x_i\|_2^2}{n^2m^3\alpha^2(1-\alpha)}\!+\!\frac{9\|\x_i\|_2^4}{n^2m(1-\alpha)}\!+\!\frac{2\|\x_i\|_2^2\|\x_i\|_1^2}{n^2m^2\alpha(1-\alpha)} \right]\\\!+\!\|\sum_{i=1}^n\frac{\|\x_i\|_1^2\x_i\x_i^2}{n^2m\alpha}\|_2$.
\end{theorem}
The proof of Theorem~\ref{the:error} is in Appendix~\ref{app:proof_error}.  

\begin{remark}
The error bound is linear with $R$ and $\sigma$.  The selected $p_{ki}$ makes $R$ and $\sigma$ smaller and tightens the bound.
\end{remark}
\begin{remark}
The balance parameter $\alpha$ can adjust the impact of the normalized $\l_1$-norm sampling and the square of the normalized $\l_2$-norm sampling. 
$\l_2$ sampling owns more potential to select larger entries to decrease error compared with $\l_1$ sampling, but $\l_2$ sampling is unstable and sensitive to small entries, incurring incredibly high estimation error if extremely small entries are picked.  Hence, if $\alpha$ varies from $1$ to $0$, the estimation error decreases first and then increases gradually.

\end{remark}

The explicit bound is represented in terms of $n$, $d$, and $m$ under the constraint of $2\leq m<d$:  
\begin{corollary} \label{cor:spec}
{Let $\X$, $m$, $\C$ and $\C_e$ be defined as in Theorem~\ref{the:unbiased}.}  Define $\frac{\|\x_i\|_1}{\|\x_i\|_2}\leq \varphi$ with $1\leq \varphi\leq\sqrt{d}$ and $\|\x_i\|_2\leq\tau$ for all $i\in [n]$.  Then, with probability at least $1-\eta-\delta$ we have  
\begin{align}
\|\C_e-\C\|_2\leq \min\{\widetilde{O}\Big(f+\frac{\tau^2\varphi}{m}\sqrt{\frac{1}{n}}+\tau^2\sqrt{\frac{1}{nm}}\Big), \notag \\ 
\widetilde{O}\Big(f+\frac{\tau\varphi}{m}\sqrt{\frac{d\|\C\|_2}{n}}+\tau\sqrt{\frac{d\|\C\|_2}{nm}}\Big)\},
\label{eq:corollary11}
\end{align}
where $f=\frac{\tau^2}{n}+\frac{\tau^2\varphi^2}{nm}+\tau\varphi\sqrt{\frac{\|\C\|_2}{nm}}$, and $\widetilde{O}(\cdot)$ hides the logarithmic factors on $\eta$, $\delta$, $m$, $n$, $d$, and $\alpha$.
\end{corollary}
The proof is given by Appendix~\ref{app:proof_spec}.

\begin{remark}\label{rem:1}
When $\varphi=\sqrt{d}$, the magnitudes of each entry in all the input data vectors are the same, i.e., highly uniformly distributed.  The error bound in Eq.~(\ref{eq:corollary11}) yields the worst case and derives a bound with a leading term of order $\min\{\widetilde{O}\Big(\frac{\tau^2d}{nm}+\tau\sqrt{\frac{d\|\C\|_2}{nm}}+\frac{\tau^2}{m}\sqrt{\frac{d}{n}}\Big), \widetilde{O}\Big(\frac{\tau^2d}{nm}+\frac{\tau d}{m}\sqrt{\frac{\|\C\|_2}{n}}\Big)\}$,  the same as \textit{Gauss-Inverse}, which ignores logarithmic factors~\cite{qi2012invariance}.  

Accordingly, as the magnitudes of the entries in each data vector becomes uneven, $\varphi$ gets smaller and yields a tighter error bound than that in \textit{Gauss-Inverse}.  Furthermore, when most of the entries in each vector $\x_i$ have very low magnitudes, the summation of these magnitudes will be comparable to a particular constant.  This situation is typical because in practice only a limited number of features in each input data dominating the learning performance.  Hence, $\varphi$ turns to $O(1)$ and Eq.~(\ref{eq:corollary11}) becomes $\min\{\widetilde{O}\Big(\frac{\tau^2}{n}+\tau^2\sqrt{\frac{1}{nm}}\Big), \widetilde{O}\Big(\frac{\tau^2}{n}+\tau\sqrt{\frac{d\|\C\|_2}{nm}}\Big)\}$, which is tighter than the leading term of \textit{Gauss-Inverse} by a factor of at least $\sqrt{d/m}$. 
\end{remark}
\begin{remark}\label{rem:2}
As practically, $m\ll d$, $O(d-m)$ approximates to $O(d)$.  The error of \textit{UniSample-HD} is  $\widetilde{O}\Big(\frac{\tau^2d}{nm}+\tau\sqrt{\frac{d\|\C\|_2}{nm}}+\frac{\tau^2d}{m}\sqrt{\frac{1}{nm}}\Big)$, which is asymptotically worse than our bound.  When $n$ is sufficiently large, the leading term of its error becomes $\widetilde{O}\Big(\tau\sqrt{\frac{d\|\C\|_2}{nm}}+\frac{\tau^2d}{m}\sqrt{\frac{1}{nm}}\Big)$, which can be weaker than the leading term in our method by a factor of $1$ to $\sqrt{d/m}$ when $\varphi=\sqrt{d}$, and at least $d/m$ when $\varphi=O(1)$. 

However, if $m$ is close to $d$, though not meaningful for practical applications, $O(d-m)=O(1)$ will hold and the error of \textit{UniSample-HD} becomes $\widetilde{O}\Big(\frac{\tau^2d}{nm}+\tau\sqrt{\frac{d\|\C\|_2}{nm}}+\frac{\tau^2}{m}\sqrt{\frac{d}{nm}}\Big)$.  This bound may slightly outperform ours by a factor of $\sqrt{d/m}=O(1)$ when $\varphi=\sqrt{d}$, but is still worse than ours when $\varphi=O(1)$.  
\end{remark}
{\bf Note.} {The derivation and proof of our DACE do not make statistical nor structural assumptions concerning the input data or the covariance matrix.}  Motivated by~\cite{azizyan2015extreme}, it is straightforward to extend our results to the data by a certain statistical assumption and a low-rank covariance matrix estimation.  
\begin{corollary} \label{cor:gau} Let $\X\in \R^{d\times n}$ ($2\leq d$), an unknown population covariance matrix $\C_p\in\R^{d\times d}$ with each column vector $\x_i\in\R^d$ i.i.d. generated from the Gaussian distribution $\mathcal{N}(\mathbf{0},\C_p)$, and $\C_e$ be constructed by Algorithm~\ref{alg:mine} with the sampling size $2\leq m<d$.  Then, with probability at least $1-\eta-\delta-\zeta$, 
\begin{align}
\frac{\|\C_e-\C_p\|_2}{\|\C_p\|_2}\leq \widetilde{O}\Big(\frac{d^2}{nm}+\frac{d}{m}\sqrt{\frac{d}{n}}\Big) \label{eq:corollary21}
\end{align}
Additionally, assuming rank$(\C_p)$$\leq r$, with probability at least $1-\eta-\delta-\zeta$, we have
\begin{align}
\frac{\|[\C_e]_r-\C_p\|_2}{\|\C_p\|_2}\leq & \widetilde{O}\Big(\frac{rd}{nm}+\frac{r}{m}\sqrt{\frac{d}{n}}+\sqrt{\frac{rd}{nm}}\Big), \label{eq:corollary22}
\end{align}
where $[\C_e]_r$ is the solution to $\min_{\text{rank}(A)\leq r}\|\A-\C_e\|_2$ and $\widetilde{O}(\cdot)$ hides the logarithmic factors on $\eta$, $\delta$, $\zeta$, $m$, $n$, $d$, and $\alpha$.
\end{corollary}

\begin{corollary} \label{cor:subspace} Given $\X$, $d$, $m$, $\C_p$ and $\C_e$ as defined in Corollary~\ref{cor:gau}.  Let $\prod_k=\sum_{i=1}^k\u_i\u_i^T$ and $\widehat{\prod}_k=\sum_{i=1}^k\hat{\u}_i\hat{\u}_i^T$ with  $\{\u_i\}_{i=1}^k$ and $\{\hat{\u}_i\}_{i=1}^k$ being the leading $k$ eigenvectors of $\C_p$ and $\C_e$, respectively. Denote by $\lambda_k$ the $k$-th largest eigenvalue of $\C_p$. Then, with probability at least $1-\eta-\delta-\zeta$, 
\begin{align}
\frac{\|\widehat{\prod}_k-\prod_k\|_2}{\|\C_p\|_2}\leq \frac{1}{\lambda_k-\lambda_{k+1}}\widetilde{O}\Big(\frac{d^2}{nm}+\frac{d}{m}\sqrt{\frac{d}{n}}\Big), \label{eq:corollary31}
\end{align}
where the eigengap $\lambda_k-\lambda_{k+1}>0$ and $\widetilde{O}(\cdot)$ hides the logarithmic factors on $\eta$, $\delta$, $\zeta$, $m$, $n$, $d$, and $\alpha$.
\end{corollary}
The sketch proof of the above two Corollaries is provided in Appendix~\ref{app:proof_corollaries}.  Corollary~\ref{cor:gau} shows that the (low-rank) covariance matrix estimation on Gaussian data and Corollary~\ref{cor:subspace} indicates that the derived covariance estimator also guarantees the accuracy of the principal components regarding the learned subspace.  In particular, setting $n=\Theta(d)$ in Eq.~(\ref{eq:corollary22}) reveals that $m=\widetilde{\Omega}(r/\epsilon^2)$ entries can achieve an $\epsilon$ spectral norm error for the low-rank covariance matrix estimation, which is also polynomial equal to the literature that leverages low-rank structure to derive methods for the low-rank (covariance) matrix recovery~\cite{cai2015rop,chen2013exact}.
\begin{table}[htbp]
\caption{Computational costs on communication (comm.) and time, where the storage of the standard method is $O(nd+d^2)$ while other methods consume $O(nm+d^2)$. 
}
\centering
\begin{tabular}{|@{~}c@{~}||c|c|c|}
\hline
Method  &  Comm. & Time\\
\hline
Standard  &  $O(nd)$ & $O(nd^2)$\\
\hline
Gauss-Inverse & $O(nm)$ & $O(nmd+nm^2d+nd^2)+T_G$\\
\hline
Sparse  & $O(nm)$ & $O(d+nm^2)+T_S$\\
\hline
UniSample-HD  & $O(nm)$ & $O(nd\log d+nm^2)$\\
\hline
Our method  & $O(nm)$ & $O(nd+nm\log d+nm^2)$ \\
\hline
\end{tabular}
\label{tab:tablemy}
\end{table}

\subsection{Computational Complexity}

In Table~\ref{tab:tablemy}, $T_G$ and $T_S$ represent the time taken by fast pseudo-random number generators like Mersenne Twister~\cite{matsumoto1998mersenne} to generate the Gaussian matrices and sparse matrices, which can be proportional to $nmd$ and $nd^2$, respectively, up to a certain small constant.  When $d$ is large, our method exhibits the most efficient method.  By applying the smallest $m$ to achieve the same estimation accuracy as the other methods, our DACE incurs the least computational cost. 


\subsection{Analysis on Multiclass Classification}
We turn to multiclass classification problem rather than image recovery~\cite{pourkamali2015estimation} because multiclass classification is popular in many real-world applications whose performance heavily depends on the estimated covariance of each individual class.  

\subsubsection{Multiclass Classification Solution}\label{subsec:mcs}
Following the setup in~\cite{jolliffe2002principal}, we are given data from $T$ classes and compute the corresponding covariance matrix for each class, denoted by $\{\C_t\}_{t=1}^T$.  Let $\prod_{k,t}=\sum_{j=1}^k\u_{j,t}\u_{j,t}^T$, where  $\{\u_{j,t}\}_{j=1}^k$ are the leading $k$ eigenvectors of $\C_t$ corresponding to the principal components.  For a test data $\x$, we fix $k$ and predict the class label by $\max_{t}\x^T\prod_{k,t}\x$.  If the mean vector in each class is zero, then the class label will be purely determined by the class covariance matrices rather than the mean vectors.

\subsubsection{Analysis} 
Let the target covariance matrix $\C_t$ in the $t$-th class be  calculated from $\{\x_{i,t}\}_{i=1}^{n_t}$, by Corollary~\ref{cor:spec}, we derive the error of our estimator $\C_{e,t}$ as follows:
\begin{align}
\frac{\|\C_{e,t}-\C_t\|_2}{\|\C_{t}\|_2}&\leq \frac{1}{\|\C_{t}\|_2}\min\{A, B\} 
\label{eq:corollary111}
\end{align}
where $A=\widetilde{O}\Big(f_t\!+\!\frac{\tau^2\varphi}{m}\sqrt{\frac{1}{n_t}}\!+\!\tau^2\sqrt{\frac{1}{n_tm}}\Big)$, $B=\widetilde{O}\Big(f_t+\frac{\tau\varphi}{m}\sqrt{\frac{d\|\C_t\|_2}{n_t}}+\tau\sqrt{\frac{d\|\C_t\|_2}{n_tm}}\Big)$, and 
$f_t=\frac{\tau^2}{n_t}+\frac{\tau^2\varphi^2}{n_tm}+\tau\varphi\sqrt{\frac{\|\C_t\|_2}{n_tm}}$, $\frac{\|\x_{i,t}\|_1}{\|\x_{i,t}\|_2}\leq \varphi$ with $1\leq \varphi\leq\sqrt{d}$, and $\|\x_{i,t}\|_2\leq\tau$ for all $i\in [n_t]$.

Compared with the estimator $\C_e$ obtained from the entire data consisting of $n=\sum_{t=1}^Tn_t$ data points, the error $\C_{e,t}$ is dominated by $1/\sqrt{n_t}$ and $1/\sqrt{\|\C_{t}\|_2}$ (not $1/\|\C_{t}\|_2$).  Suppose $n_t$ is the same for all classes, the term $1/\sqrt{n_t}$ in Eq.~(\ref{eq:corollary111}) becomes $\sqrt{T/n}$, which is $\sqrt{T}$ times as large as $1/\sqrt{n}$ in Eq.~(\ref{eq:corollary11}) of Corollary~\ref{cor:spec}.  Meanwhile, if all $\C_{t}$ have very similar principal components as those of $\C$ (i.e., all $\{\C_{t}\}_{t=1}^T$ and $\C$ are calculated from the same data distributions and thus the data tend to be difficult to classify), then $1/\sqrt{\|\C_{t}\|_2}$ nearly equals $1/\sqrt{\|\C\|_2}$.  If all $\C_{t}$ have totally different principal components from each other, $1/\sqrt{\|\C_{t}\|_2}$ will become $1/\sqrt{T}$ times as large as $1/\sqrt{\|\C\|_2}$. 

Thus, compared with the estimation error derived in Corollary~\ref{cor:spec} over $n$ data with $T$ classes, the estimation error for each class covariance estimator roughly increases with $O(\sqrt{T})$ if all class covariance matrices yield nearly the same leading principal components while the error remaining nearly the same if all class covariance matrices have totally different leading principal components. 

\begin{remark}
The above result does not contradict with our statement in Theorem~\ref{the:error} or Corollary~\ref{cor:spec} that the estimation error approaches zero (i.e., decreases with the number of data) when receiving more data, because the data follow only one category of distribution result in a stable $1/\sqrt{\|\C\|_2}$.  When the data follows different distributions, they practically come randomly and yield a stable $1/\sqrt{\|\C\|_2}$. 
\end{remark}

\begin{remark}
In Sec.~\ref{subsec:mcs}, the performance is determined only by $\C_{e,t}$ or the principal components $\prod_{k, t}$ of $\C_{e,t}$.  Via Corollary~\ref{cor:spec} and Corollary~\ref{cor:subspace}, the compressed data obtained by our method guarantee a superior approximation for $\C_{e,t}$ and $\prod_{k,t}$ and yields a comparable classification performance compared with $\C_{t}$. In other words, if each $n_t$ or the target compressed dimension $m$ is not too small, the distances between individual class estimators will approach to those among the original class covariance matrices with high probability and guarantee the classification performance.
\end{remark}

\begin{figure}[htbp]
\centering
\subfigure{
\includegraphics[width=.22\textwidth]{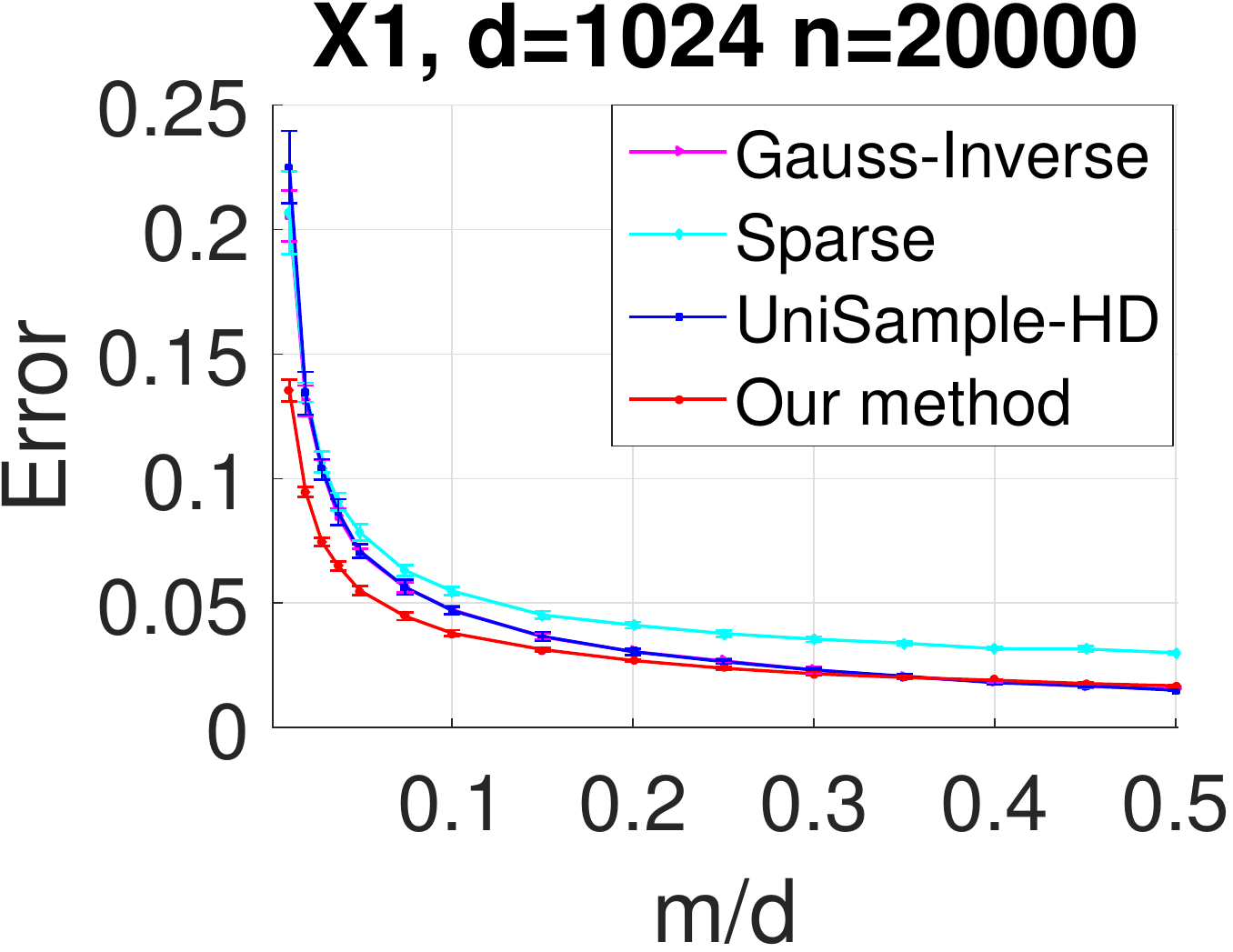}} 
\subfigure{
\includegraphics[width=.22\textwidth]{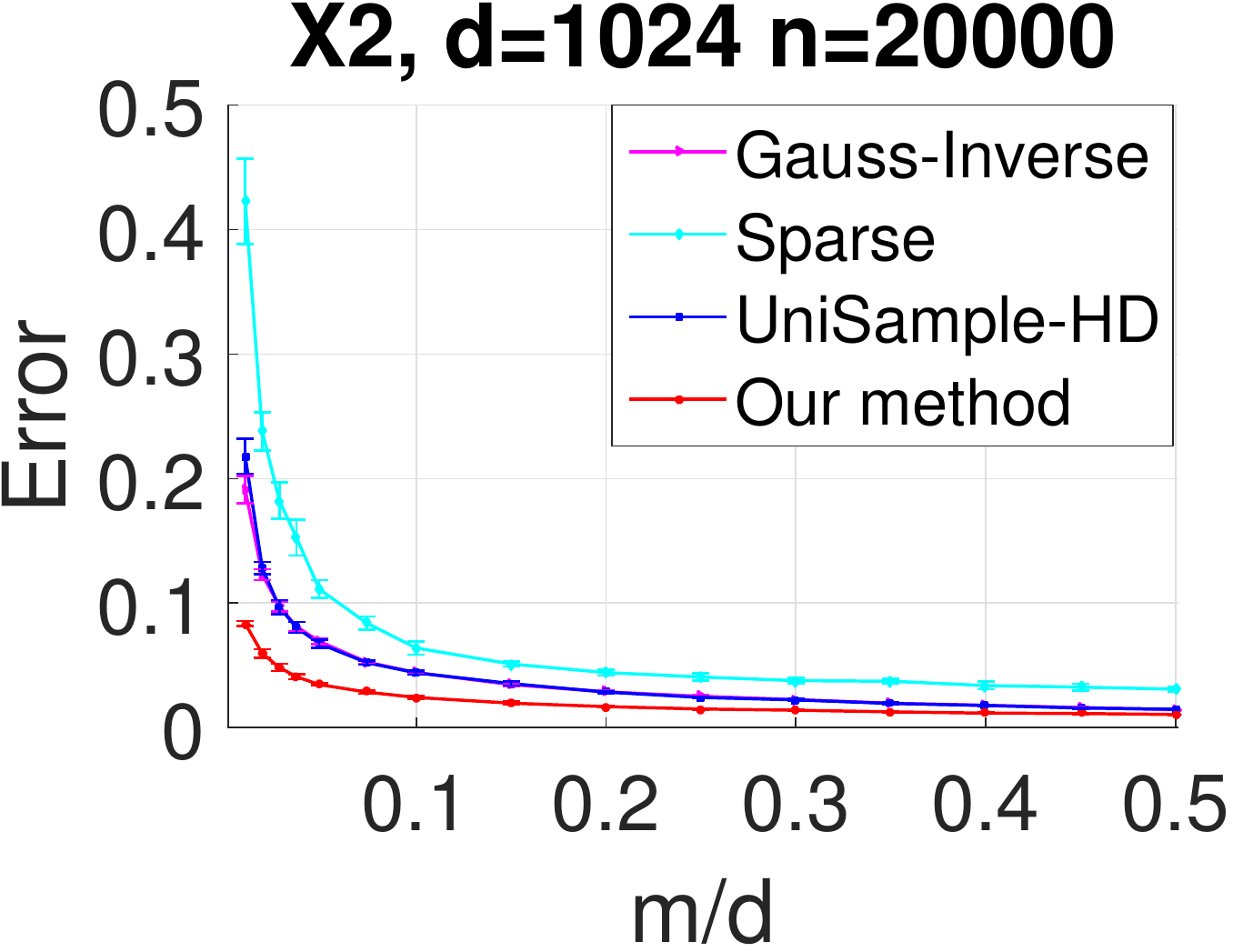}}
\subfigure{
\includegraphics[width=.22\textwidth]{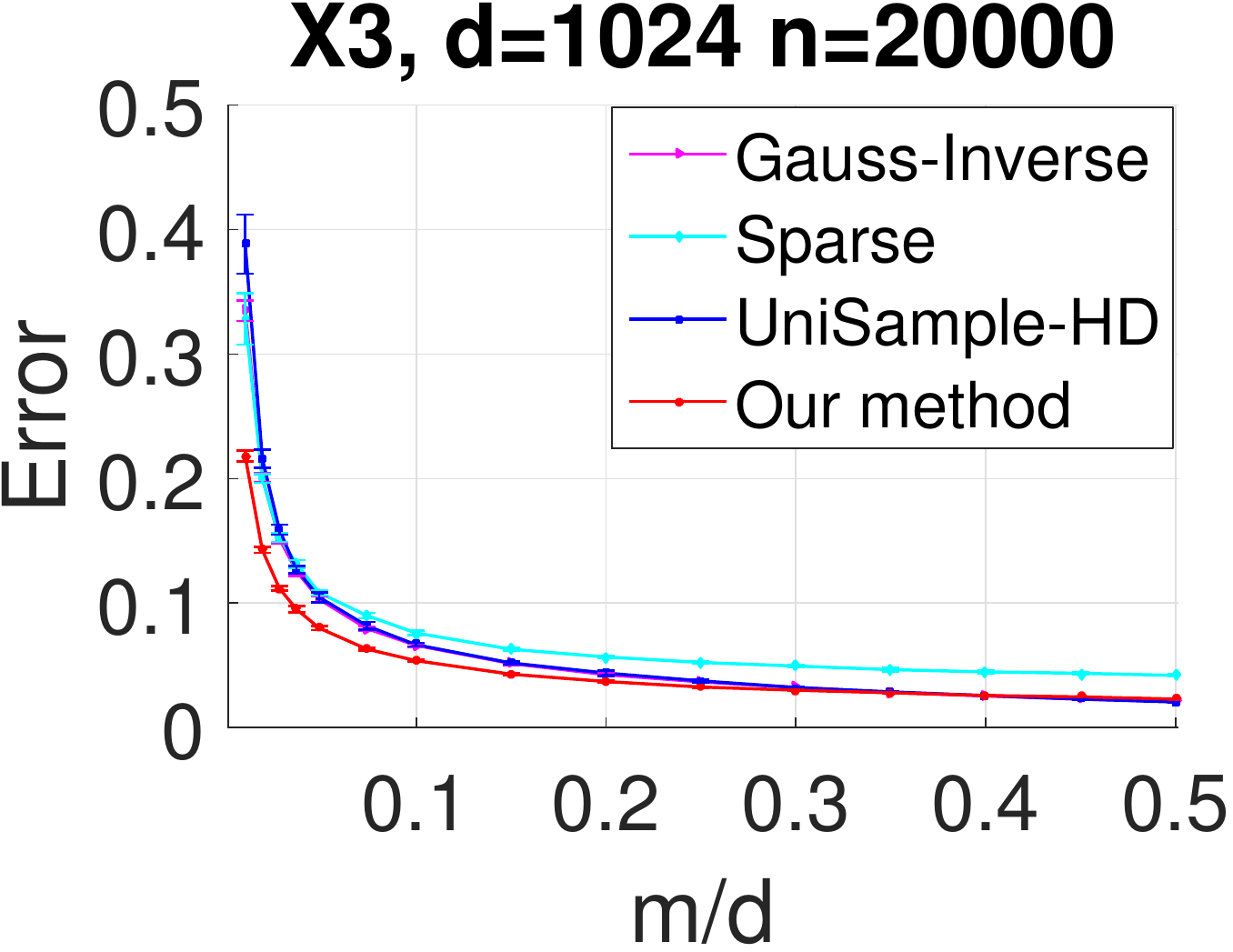}}
\subfigure{
\includegraphics[width=.22\textwidth]{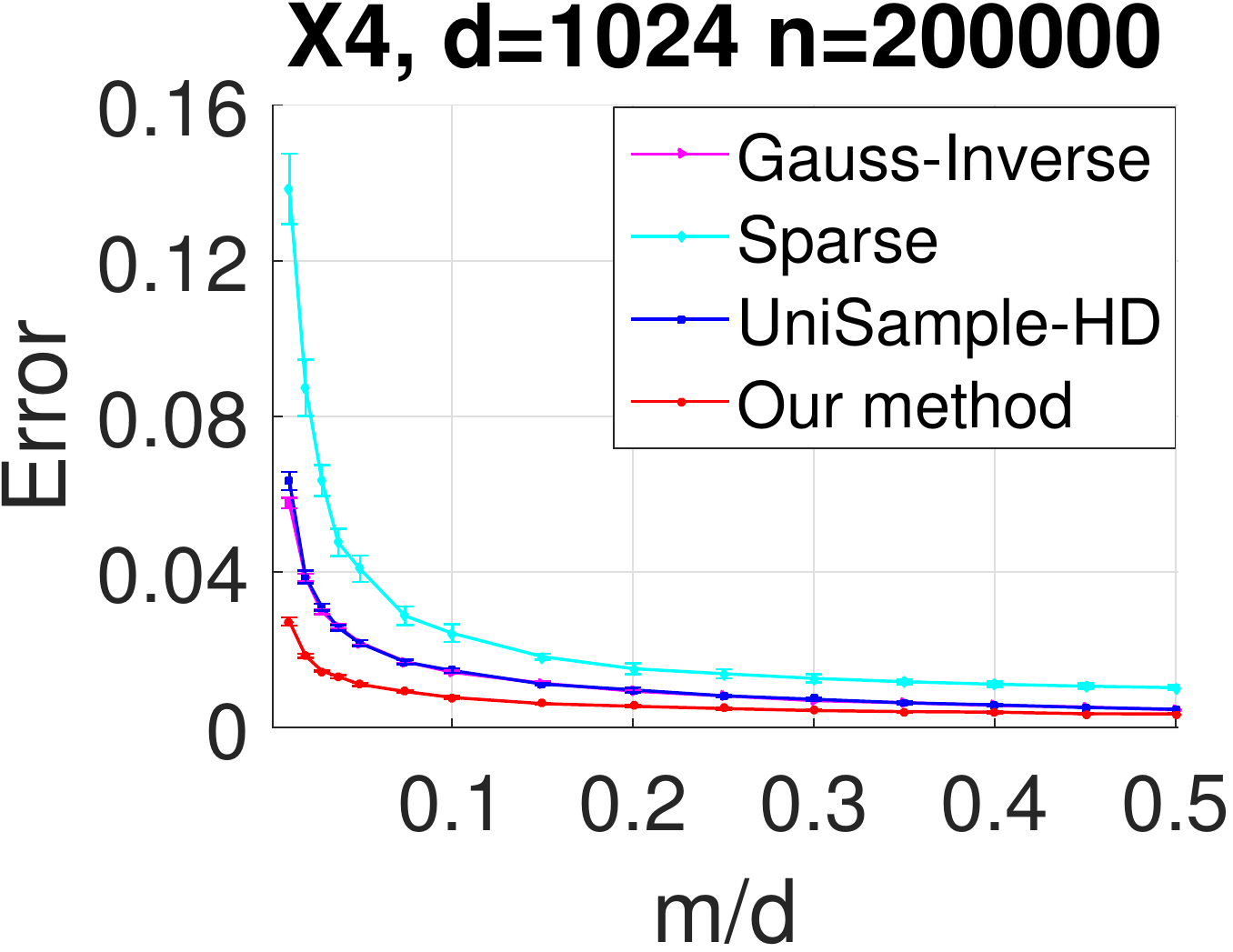}}
\subfigure{
\includegraphics[width=.22\textwidth]{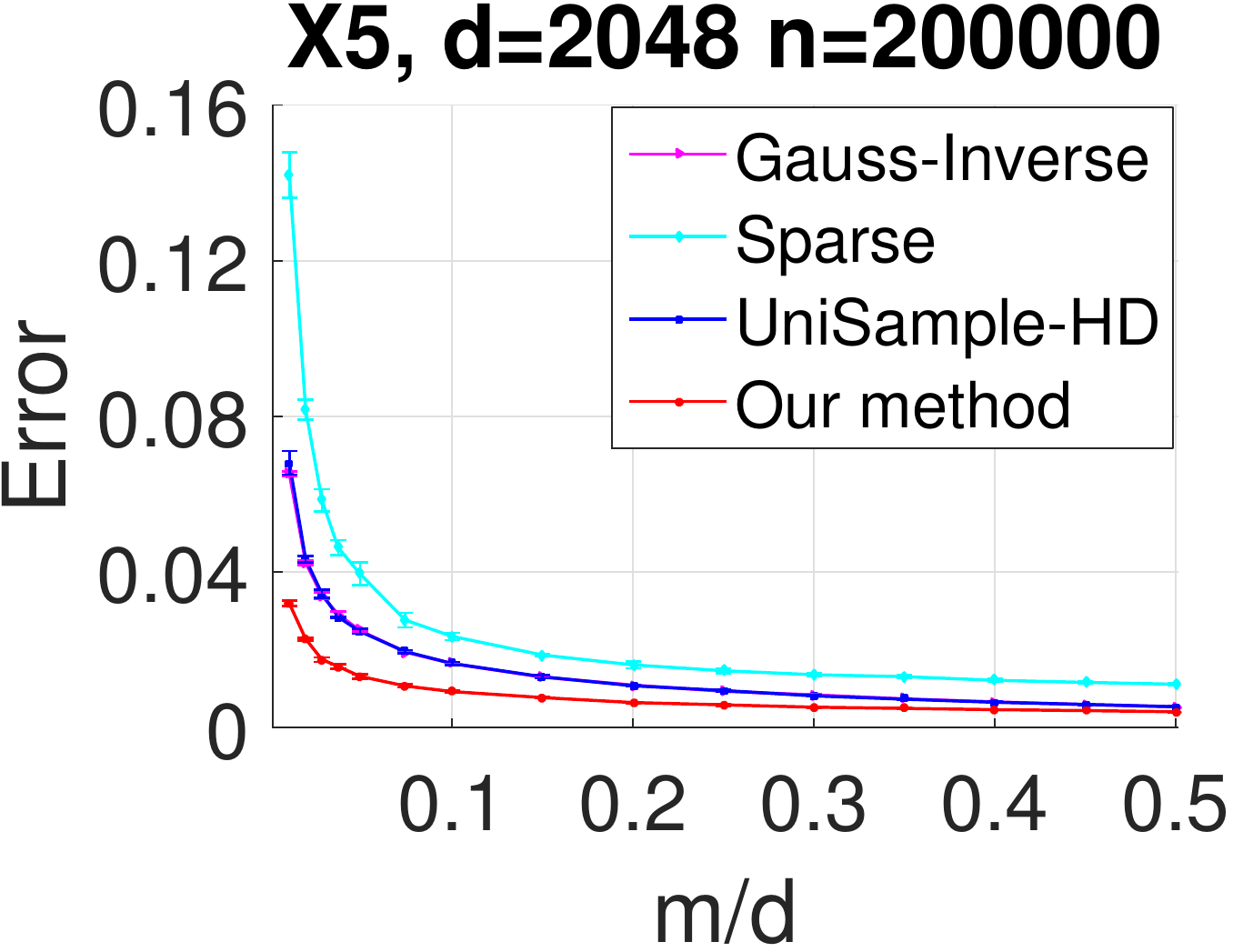}}
\subfigure{
\includegraphics[width=.22\textwidth]{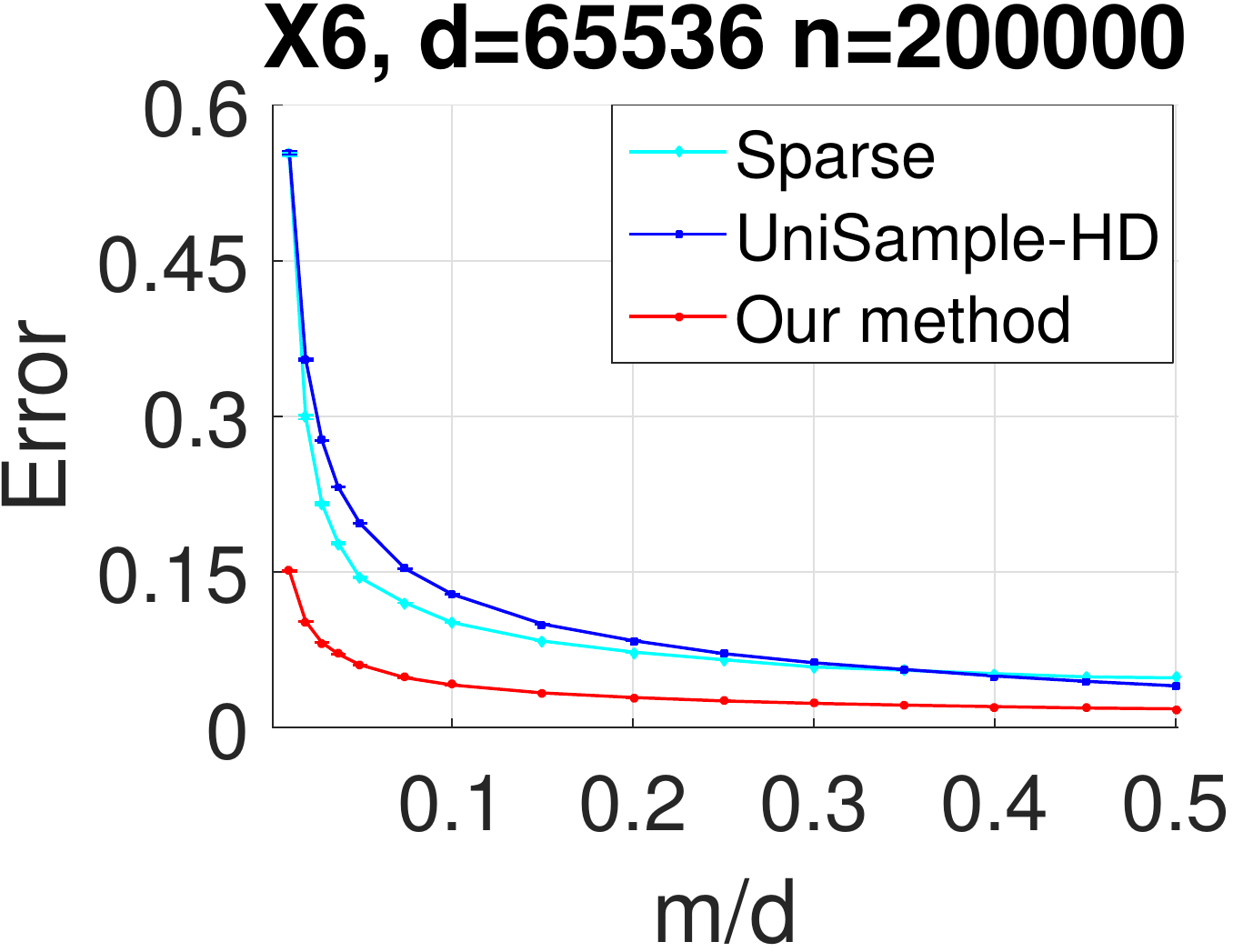}}
\caption{Accuracy comparisons of covariance matrix estimation on synthetic datasets.  The estimation error is measured by $\|\C_e-\C\|_2/\|\C\|_2$ with $\C_e$ calculated by all compared methods and $\mathrm{cf}=m/d$ is the compression ratio.}
\label{fig:accuracysyn}
\end{figure}
\section{Experiments} \label{sec:exp}
{
In this section, we conduct empirical evaluation to address the following issues:
\begin{compactitem}
\item {How the dimension, the data size, and the compression ratio affect the estimation precision of our DACE?}
\item {What performance of our DACE can be attained in real-world datasets?}
\item {What performance of our DACE can be attained in multiclass classification problems?}
\end{compactitem}

To provide fair comparisons, we compare our DACE with three representative algorithms: \textit{Gauss-Inverse}~\cite{qi2012invariance}, \textit{Sparse}~\cite{anaraki2014memory}, and \textit{UniSample-HD}~\cite{pourkamali2015preconditioned}.
}
{
In our DACE, the hyper-parameter $\alpha$ is empirically set to $0.9$ due to good empirical performance.  The hyper-parameter settings in \textit{Gauss-Inverse}, \textit{Sparse}, and \textit{UniSample-HD} simply follow the original work in~\cite{pourkamali2015preconditioned,qi2012invariance,anaraki2014memory,qi2012invariance,anaraki2014memory}.
}

All algorithms are implemented in C++ and run in a single thread mode on a standard workstation with Intel CPU@2.90GHz and $128$GB RAM to record the time consumption measured by FLOPS.

\subsection{Covariance Estimation on Synthetic Datasets}
Following the generation procedure~\cite{liberty2013simple}, we construct six synthetic datasets: 1) $\{\X_i\}_{i=1}^3$, $d=\num{1024}$, $n=\num{20000}$; 2) $\X_4$, $d=\num{1024}$, $n=\num{200000}$; 3) $\X_5$, $d=\num{2048}$, $\num{200000}$, and $\X_6$, $d=\num{65536}$, $n=\num{200000}$.  More specifically, $\X=\U\F\G$, where $\U\in\R^{d\times k}$ ($\U^T\U=\I_k$, $k\leq d$) defines the signal column space, the square diagonal matrix $\F\in\R^{k\times k}$ contains the diagonal entries $f_{ii}=1-(i-1)/k$ with linearly diminishing signal singular values, and $\G\in\R^{k\times n}$ is the Gaussian signal, i.e., $g_{ij}\sim\mathcal{N}(0,1)$.  In $\X_1$, $k\approx 0.005d$.   $\X_2=\D\X$, where $\D$ is a square diagonal matrix with $d_{ii}=1/\beta_i$ and integer $\beta_i$ is uniformly sampled from range $1$-$15$.  $\X_3$ is constructed the same way as $\X_1$ except that $\F$ is set by an identity matrix.  $\{\X_i\}_{i=4}^6$ follow the same generation scheme of $\X_2$ by only setting different $n$ and $d$. 

\begin{figure}[htbp]
\centering
\subfigure{
\includegraphics[width=.15\textwidth]{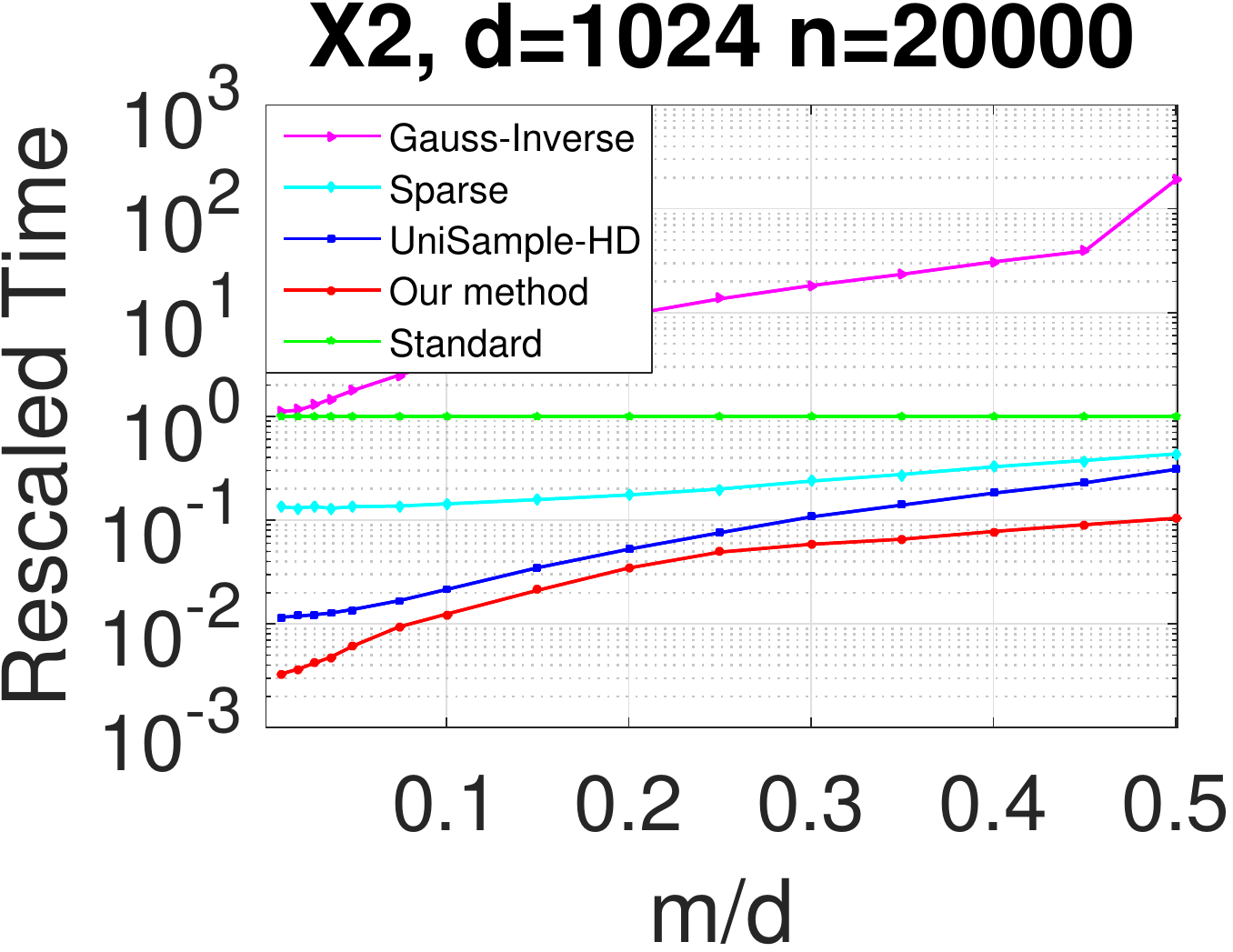} 
}
\hspace{-0.115in}
\subfigure{

\includegraphics[width=.15\textwidth]{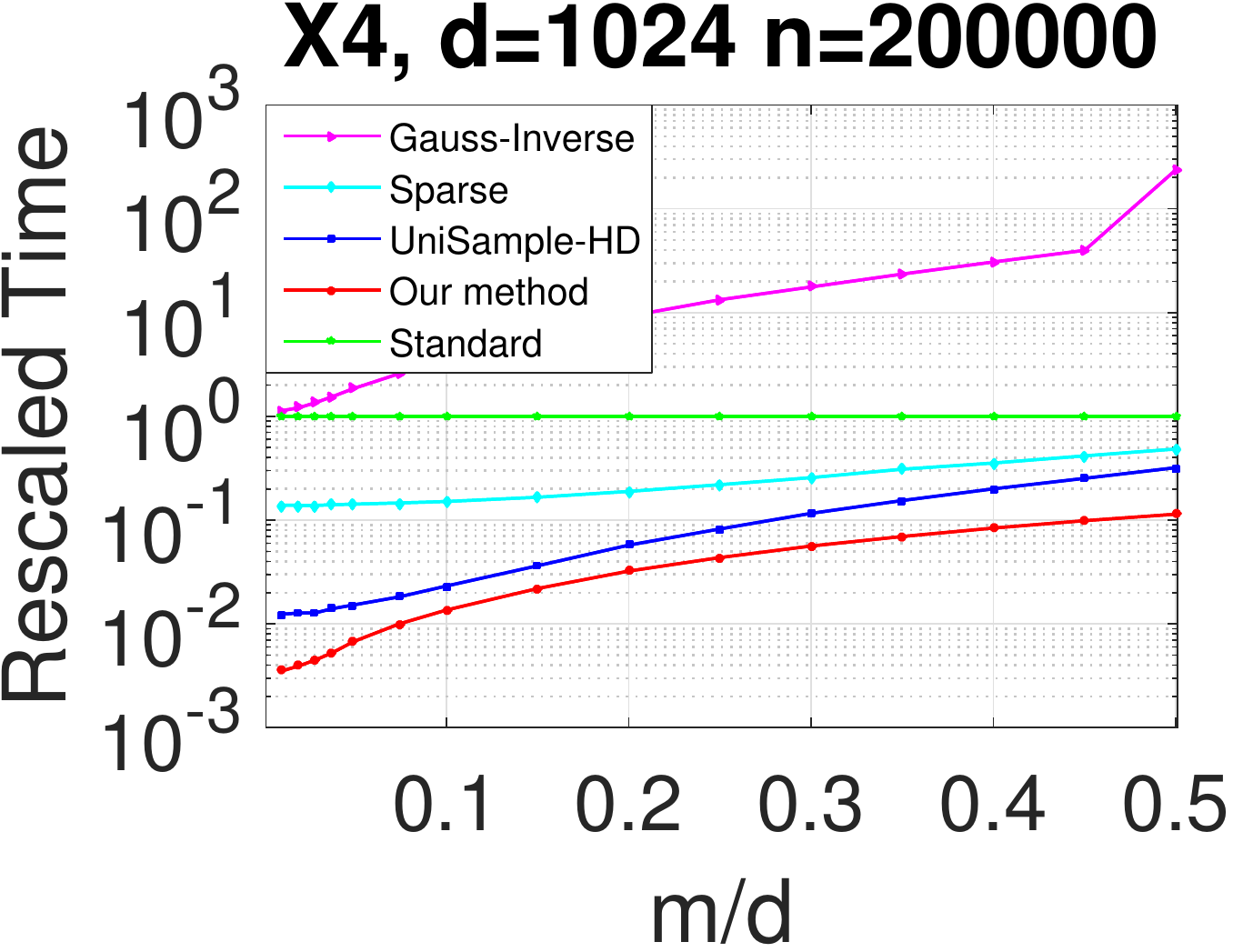}
}
\hspace{-0.115in}
\subfigure{
\includegraphics[width=.15\textwidth]{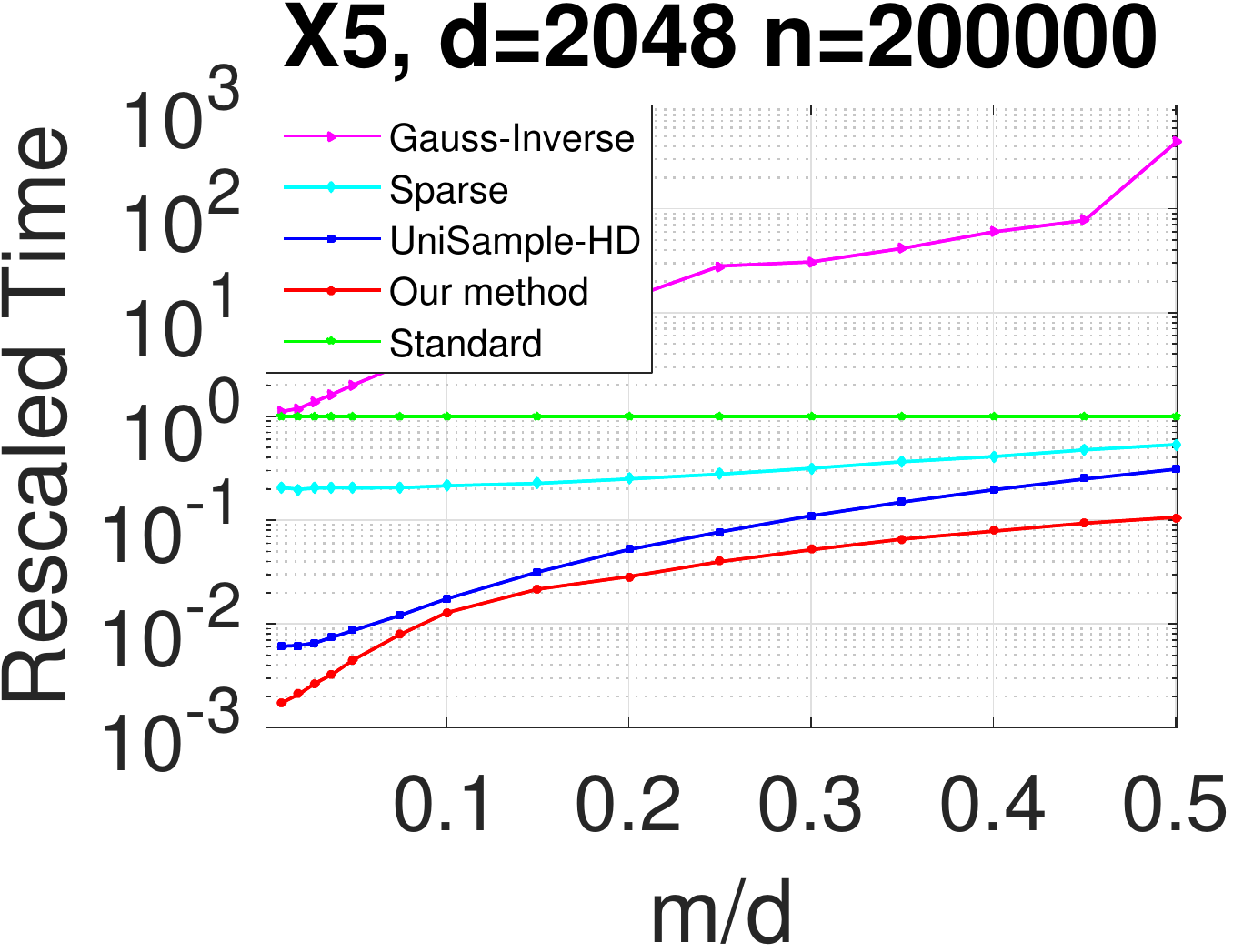}
}
\caption{Rescaleed time cost (plotted in the log scale) of covariance matrix estimation on synthetic datasets.  The time is normalized to the \textit{Standard} way of calculating $\C=\X\X^T/n$ on the original data.}
\label{fig:timesyn}
\end{figure}

Figure~\ref{fig:accuracysyn} plots the average relative estimation error with its standard deviation on ten runs with respect to the compression ratio $\mathrm{cf}=m/d$.  Note that the performance of \textit{Gauss-Inverse} has been revealed on $\X_1$ to $\X_5$ and is not provided on $\X_6$ due to enormous computation time.  Figure~\ref{fig:timesyn} reports the rescaled time cost in both the compressing and recovering stages.  The results show that
\begin{compactitem}
\item Our DACE exhibits the least estimation error and deviation for all datasets when the dimension $d$ increases.  When applying more data, the error decreases gradually.  The error decreases dramatically with slightly increasing $\mathrm{cf}$ and becomes flat soon.  It indicates that our DACE can achieve sufficient estimation accuracy by using substantially fewer data entries than other methods. 
\item When the compression ratio $\mathrm{cf}$ increases, the estimation error decreases gradually while the time cost increases accordingly.  \textit{Gauss-Inverse}, though good for smaller storage and less communication, consumes significantly much more time than \textit{Standard} due to the computation of non-sparse projection.  \textit{Sparse}, though without error analysis of the estimator, generally consumes less time than \textit{Standard}, but performs worse than the other methods.  \textit{UniSample-HD} beats other two methods, but performs slightly worse than our DACE and consumes more time than DACE.  Especially, the inferiority is remarkable when $\mathrm{cf}$ is small.   
\item In $\X_1$, $\varphi$ is measured empirically as $0.81\sqrt{d}$, the magnitudes of the data entries are more uniformly distributed, our DACE can be regarded as uniform sampling with replacement and may perform slightly worse than \textit{UniSample-HD} and \textit{Gauss-Inverse}.  {In $\X_2$, $\varphi=0.55\sqrt{d}$, the magnitude varies in a moderately larger range, our DACE outperforms the three other methods significantly.  The improvement lies that our DACE is \textit{only} sensitive to $\varphi$ and a smaller $\varphi$ produces a tighter estimation, which confirms the elaboration in Remark~\ref{rem:1} and Remark~\ref{rem:2}. } 
\item The error of each method in $\X_3$ ($\tau/\sqrt{\|\C\|_2}=5.5, \varphi=0.81\sqrt{d}$) is larger than that in $\X_1$ ($\tau/\sqrt{\|\C\|_2}=4.3, \varphi=0.81\sqrt{d}$), respectively. It is because of that almost \textit{all} methods are sensitive to $\tau/\sqrt{\|\C\|_2}$, and the error $\|\C_e-\C\|_2/\|\C\|_2$ increases when $\tau/\sqrt{\|\C\|_2}$ increases.  Such phenomenon is demonstrated via dividing numerous error bounds in Remarks~\ref{rem:1} and~\ref{rem:2} by $\|\C\|_2$.  Our method also achieves the best performance in $\X_4$. Although the $\varphi$ and $\tau/\sqrt{\|\C\|_2}$ in $\X_4$ are approximately equal with those in $\X_2$, yet the proved error bounds with Remarks~\ref{rem:1} and~\ref{rem:2} reveal that a larger $n$ in $\X_4$ will lead to smaller estimation errors given the same $\mathrm{cf}$.
\end{compactitem}

\begin{figure}[htbp]
\centering
\subfigure[]{\label{fig:error_n}
\includegraphics[width=.22\textwidth]{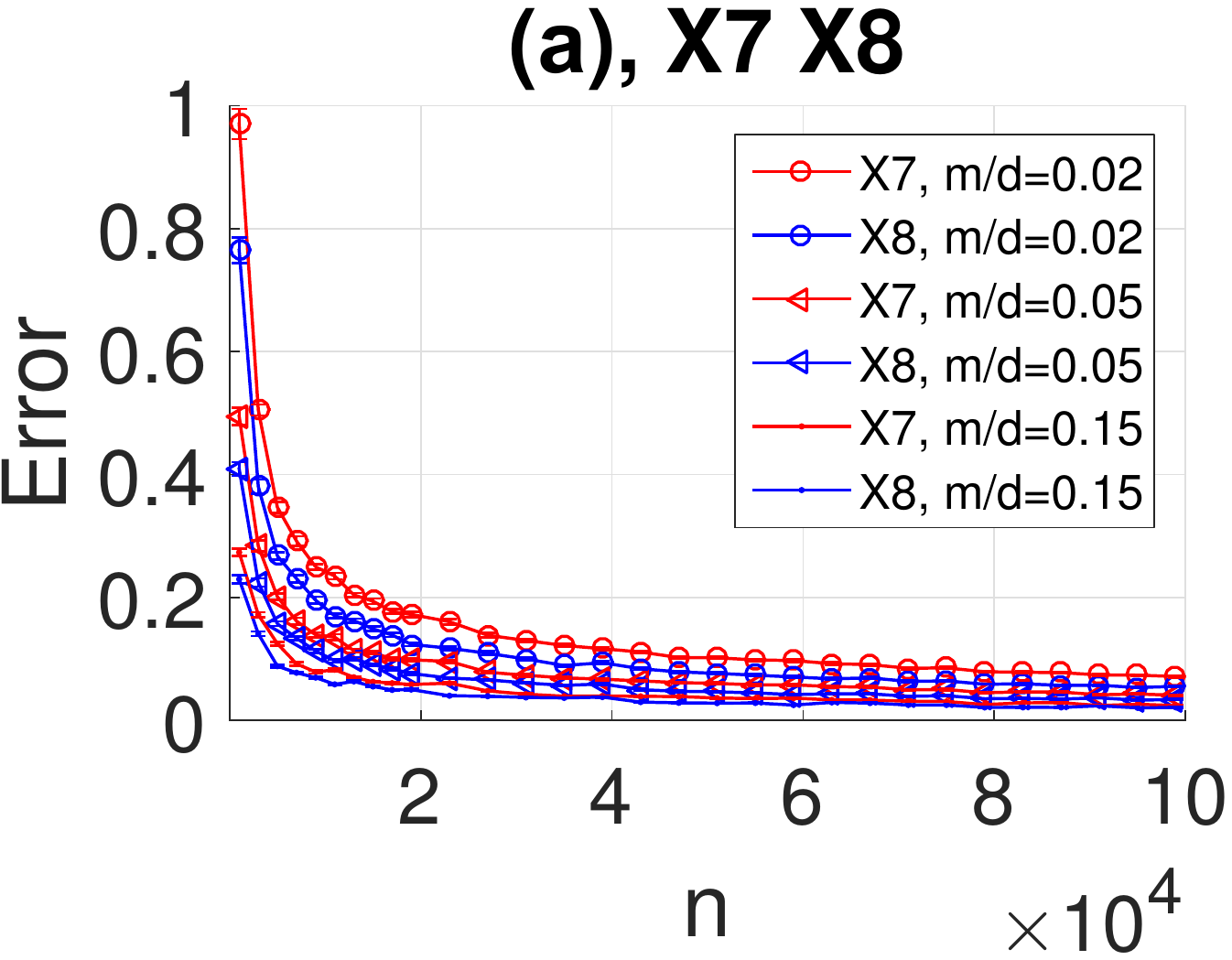}
}
\subfigure[]{\label{fig:rel_error_n}
\includegraphics[width=.22\textwidth]{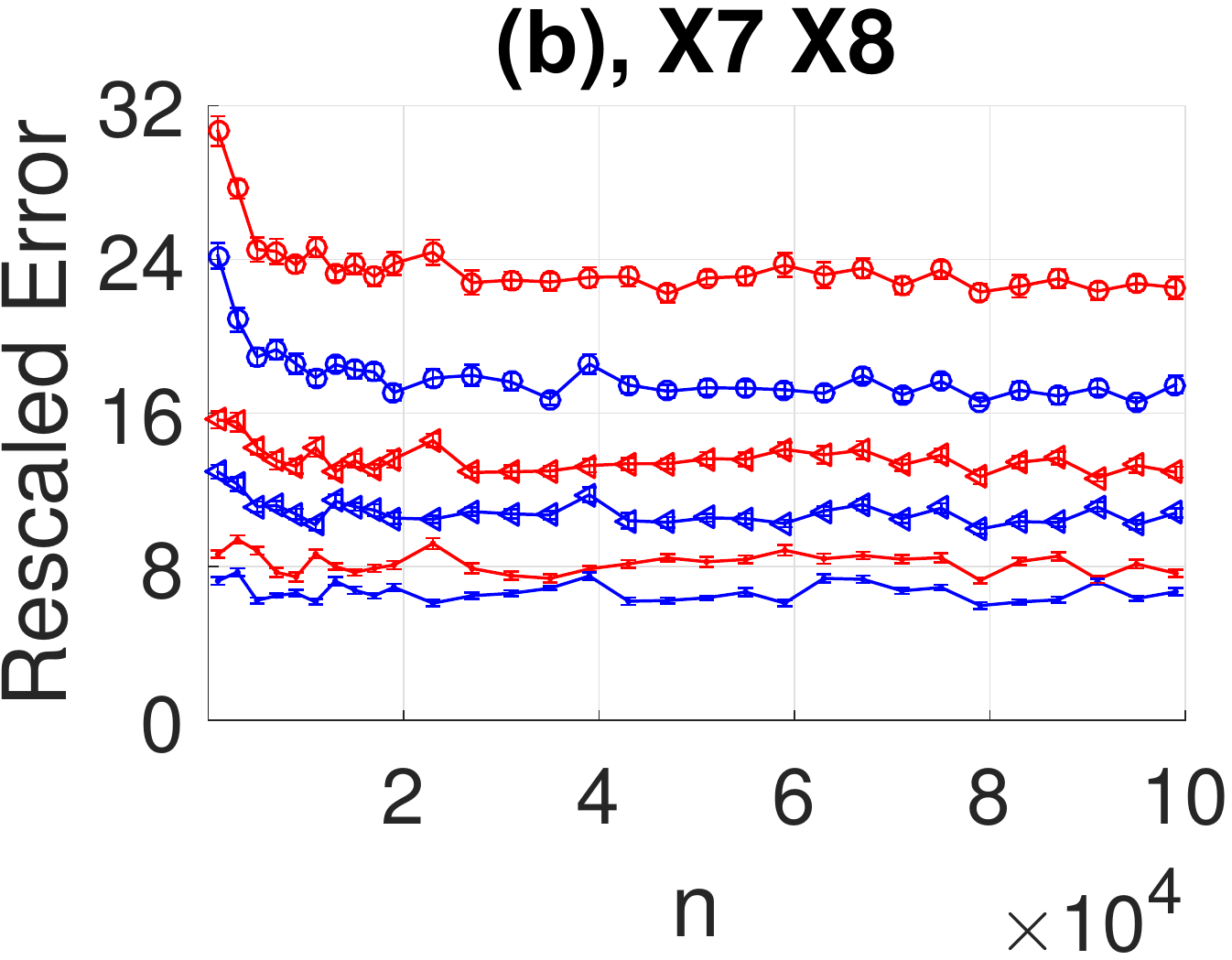}
}
\subfigure[]{\label{fig:error_X8_n}
\includegraphics[width=.22\textwidth]{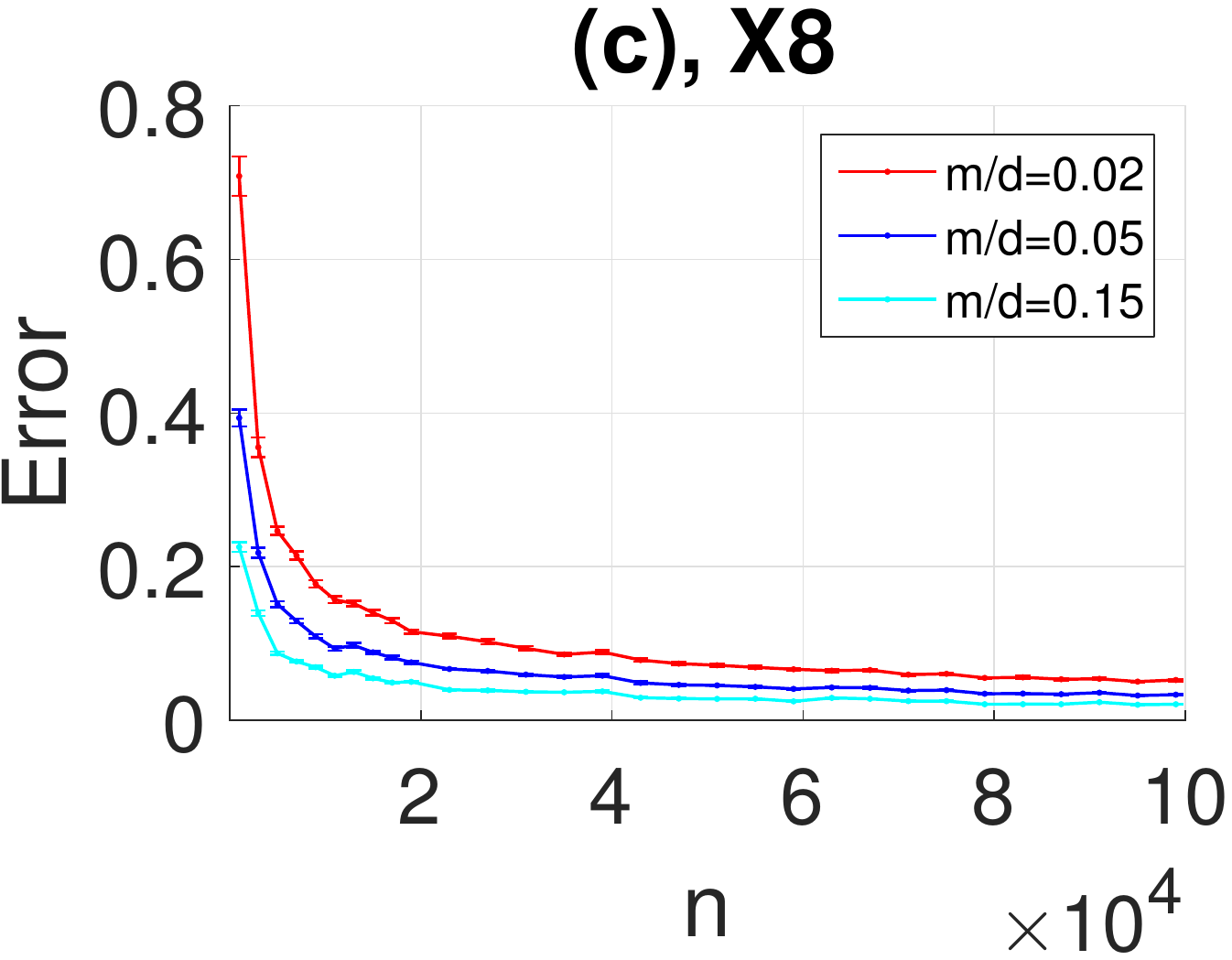}
}
\subfigure[]{\label{fig:rel_error_X8_n}
\includegraphics[width=.22\textwidth]{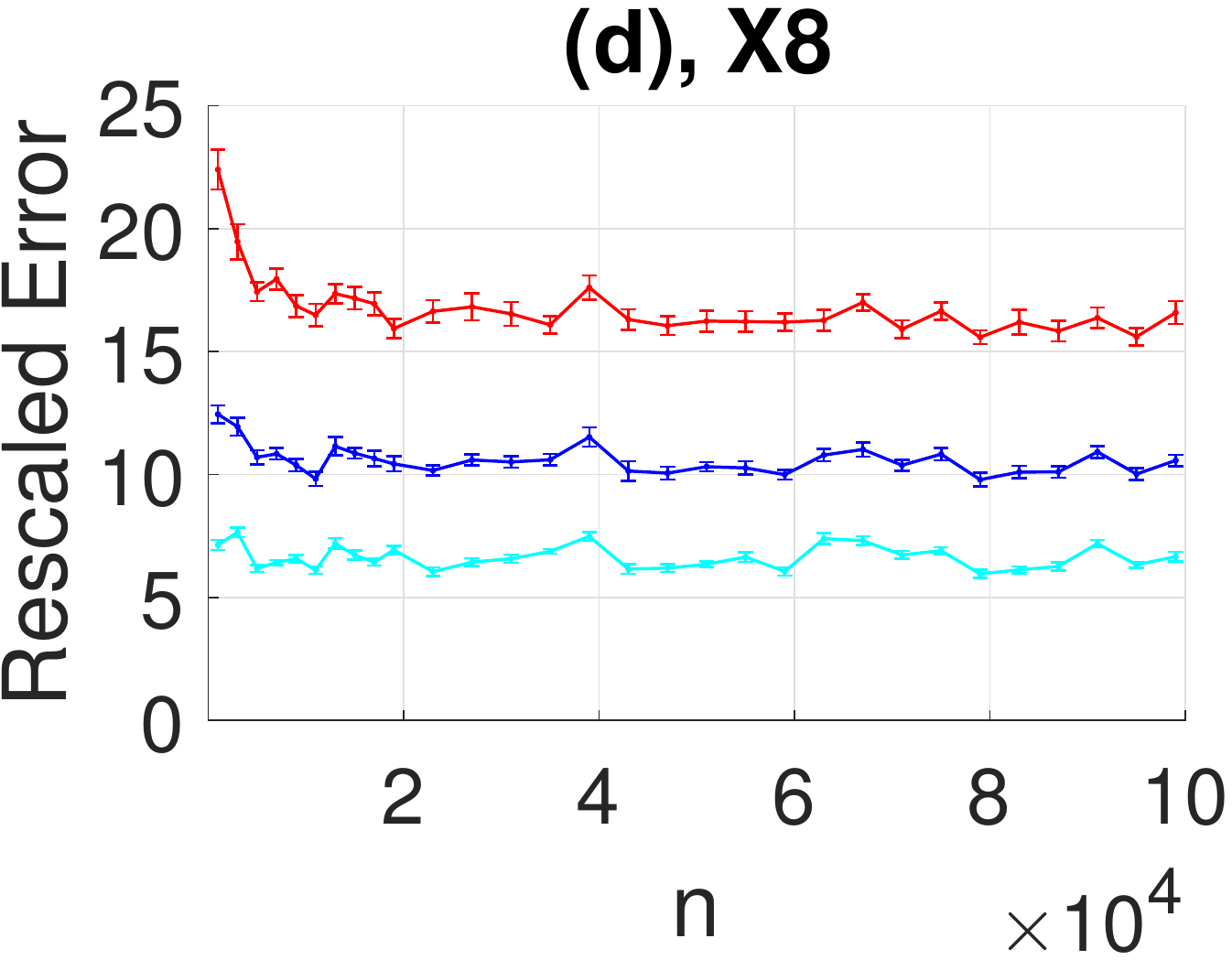}}
\subfigure[]{\label{fig:error_X7_n}
\includegraphics[width=.22\textwidth]{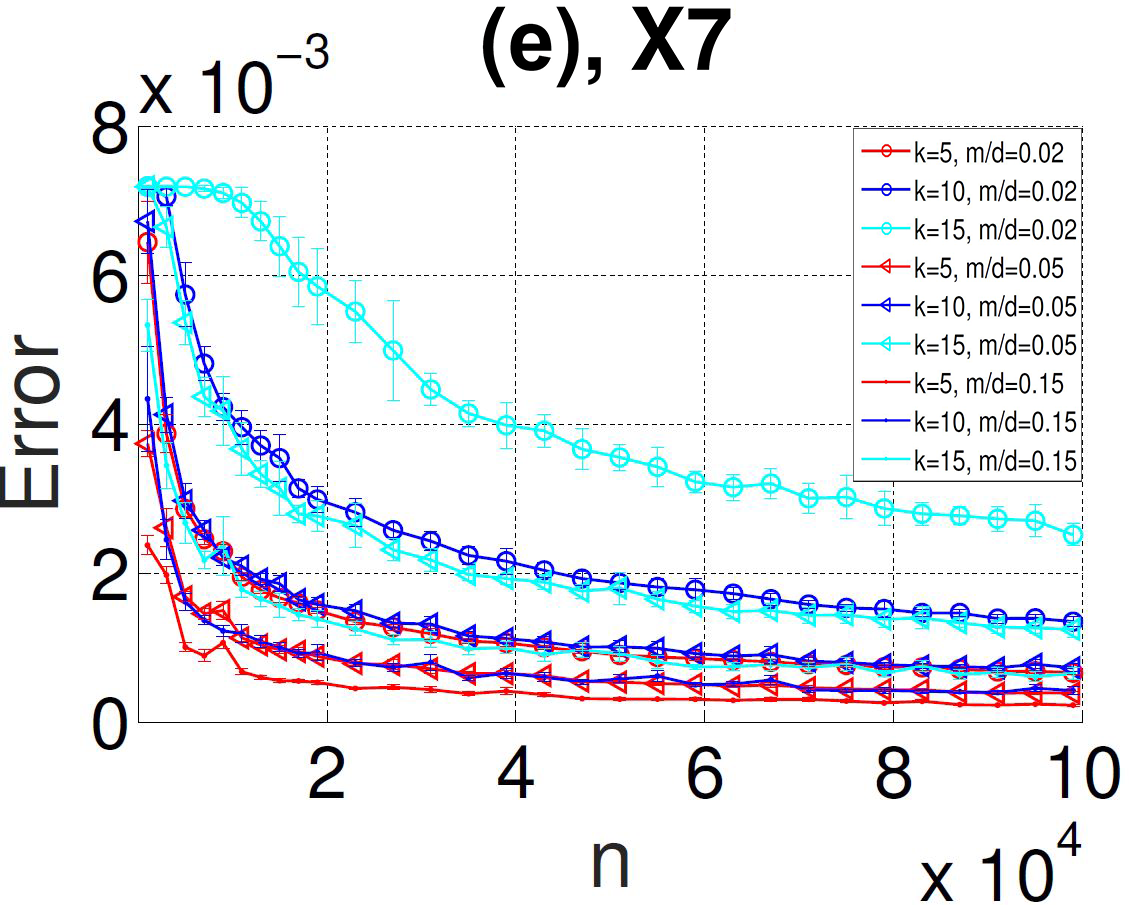}}
\subfigure[]{\label{fig:rel_error_X7_n}
\includegraphics[width=.22\textwidth]{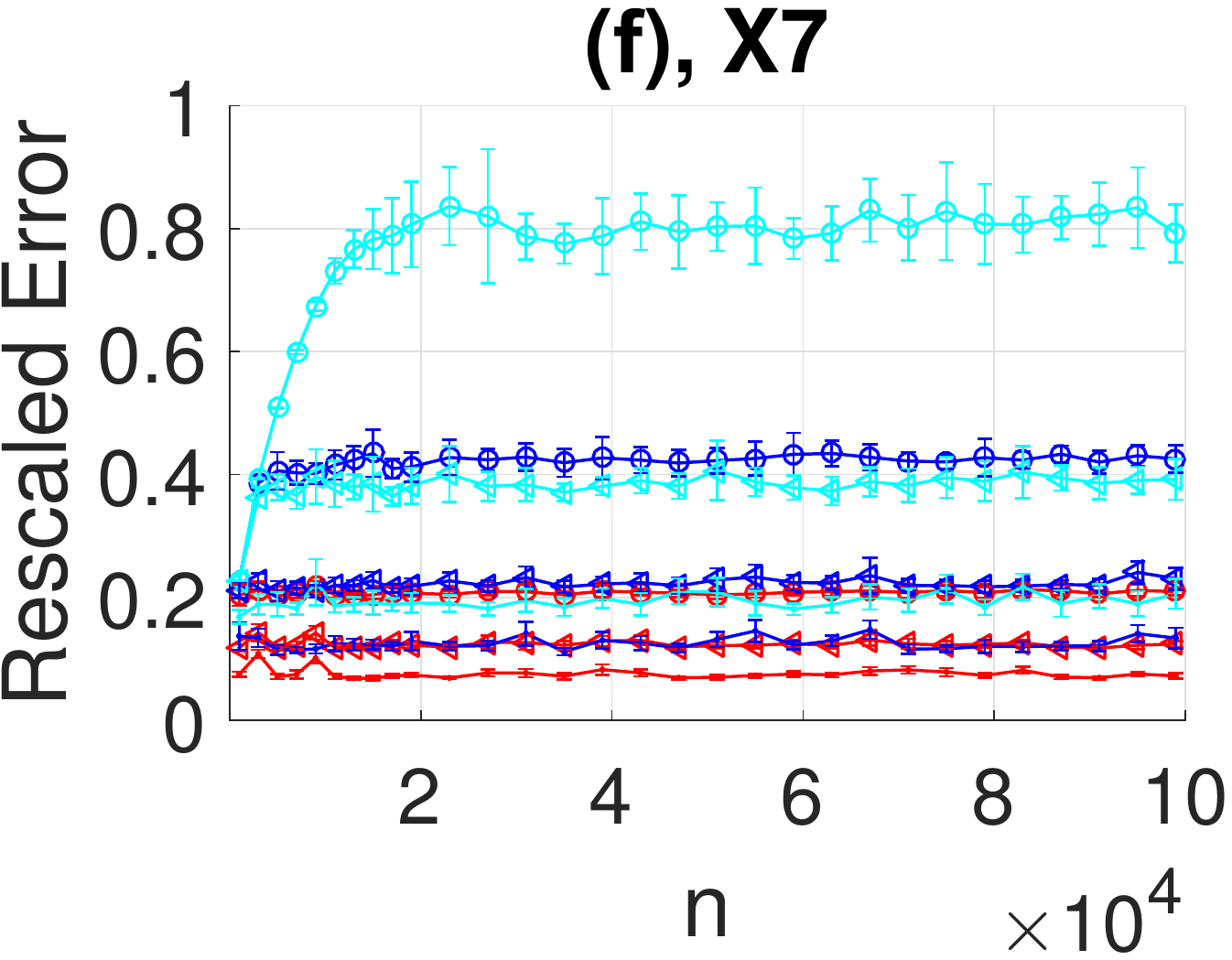}}
\caption{Convergence analysis of CACE in terms of different $n$,  $\mathrm{cf}$, and $k$.}
\label{fig:addsyn}
\end{figure}

To verify the theoretical results in Corollaries~\ref{cor:gau}-\ref{cor:subspace}, we generate two new synthetic datasets from multivariate normal distribution $\{\X_t\sim\mathcal{N}(\mathbf{0},\C_{pt})\}_{t=7}^8\in\R^{d\times d}$.  In $\X_7$, the $(i, j)$-th element of $\C_{p7}$ is $0.5^{|i-j|/{50}}$ while $\C_{p8}$ being a low-rank matrix which is the solution to $\min_{\mbox{rank}(\A)\leq r}\|\A-\C_{p7}\|_2$.  We take $d=\num{1000}$, $r=5$, $m/d=\{0.02, 0.05, 0.15\}$, $k=\{5, 10, 15\}$, and vary $n$ from $\num{1000}$ to $\num{100000}$.  

Fig.~\ref{fig:error_n}, Fig.~\ref{fig:error_X8_n}, and Fig.~\ref{fig:error_X7_n} report the errors defined in the LHS of Eqs.~(\ref{eq:corollary21})-(\ref{eq:corollary31}) under different settings while Fig.~\ref{fig:rel_error_n}, Fig.~\ref{fig:rel_error_X8_n}, and Fig.~\ref{fig:rel_error_X7_n} recording the errors divided by $1/\sqrt{n}$.  The results show that the errors decreases as the increases of $n$.  Especially, the roughly flat curves in Fig.~\ref{fig:rel_error_n}, Fig.~\ref{fig:rel_error_X8_n}, and Fig.~\ref{fig:rel_error_X7_n} indicate that the error bounds induced by our DACE converge rapidly in the rate of $1/\sqrt{n}$, which coincides with the results in Eqs.~(\ref{eq:corollary21})-(\ref{eq:corollary31}).  Fig.~\ref{fig:error_n} also exhibits that our DACE can attain more accurate estimation precision from a low-rank generated covariance matrix  than that from a high-rank covariance matrix, and enlarging $n$ can improve all the estimation precision. Fig.~\ref{fig:error_X7_n} shows that the estimated errors increase with the increase of $k$, and these results cohere with Eq.~(\ref{eq:corollary31}) by considering the empirical findings that the eigengap $\lambda_k-\lambda_{k+1}$ in $\C_{p7}$ decreases with $k$.

\begin{figure}[htbp]
\centering
\subfigure{
\includegraphics[width=.22\textwidth]{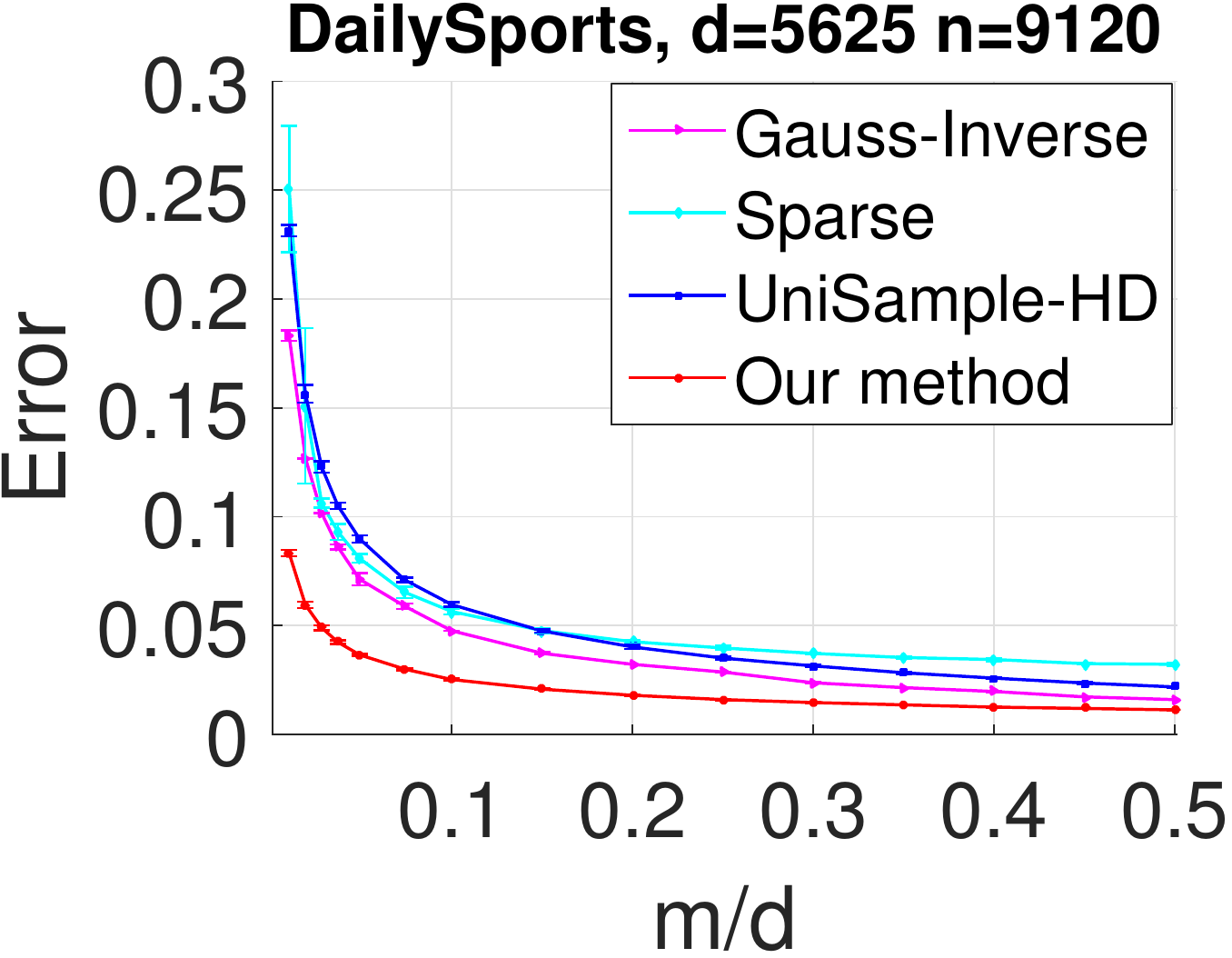}
}
\subfigure{
\includegraphics[width=.22\textwidth]{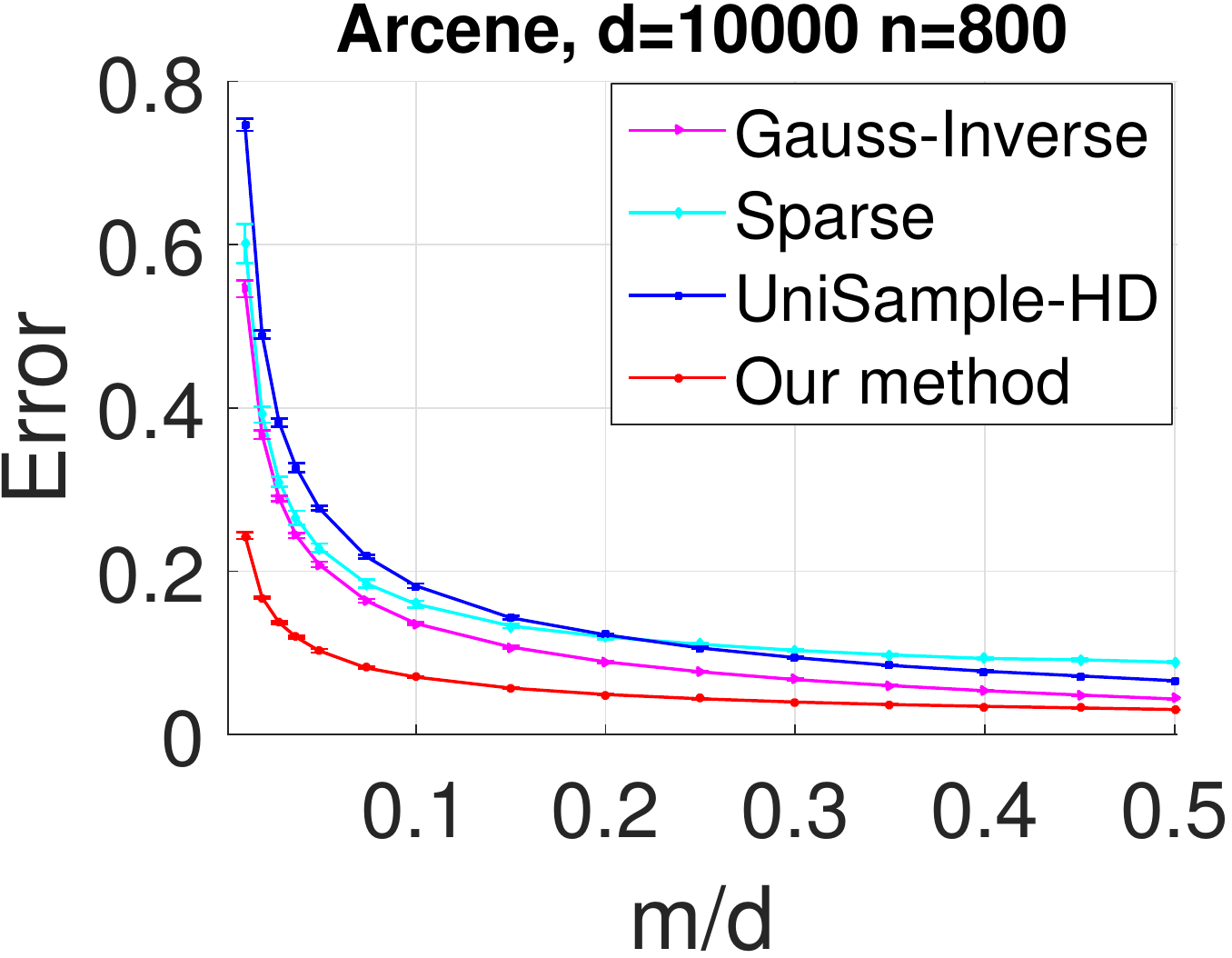}
}
\subfigure{
\includegraphics[width=.22\textwidth]{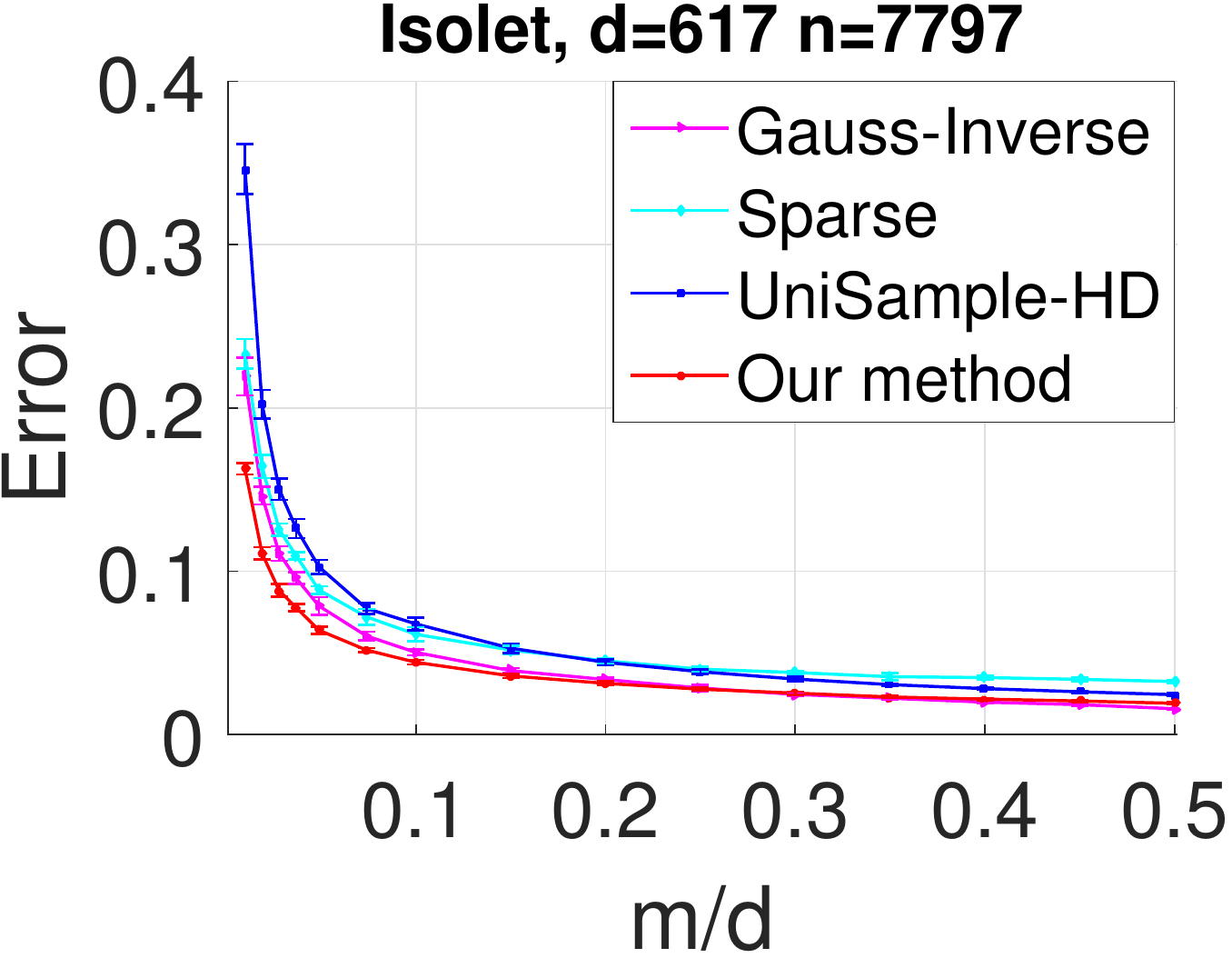}
}
\subfigure{
\includegraphics[width=.22\textwidth]{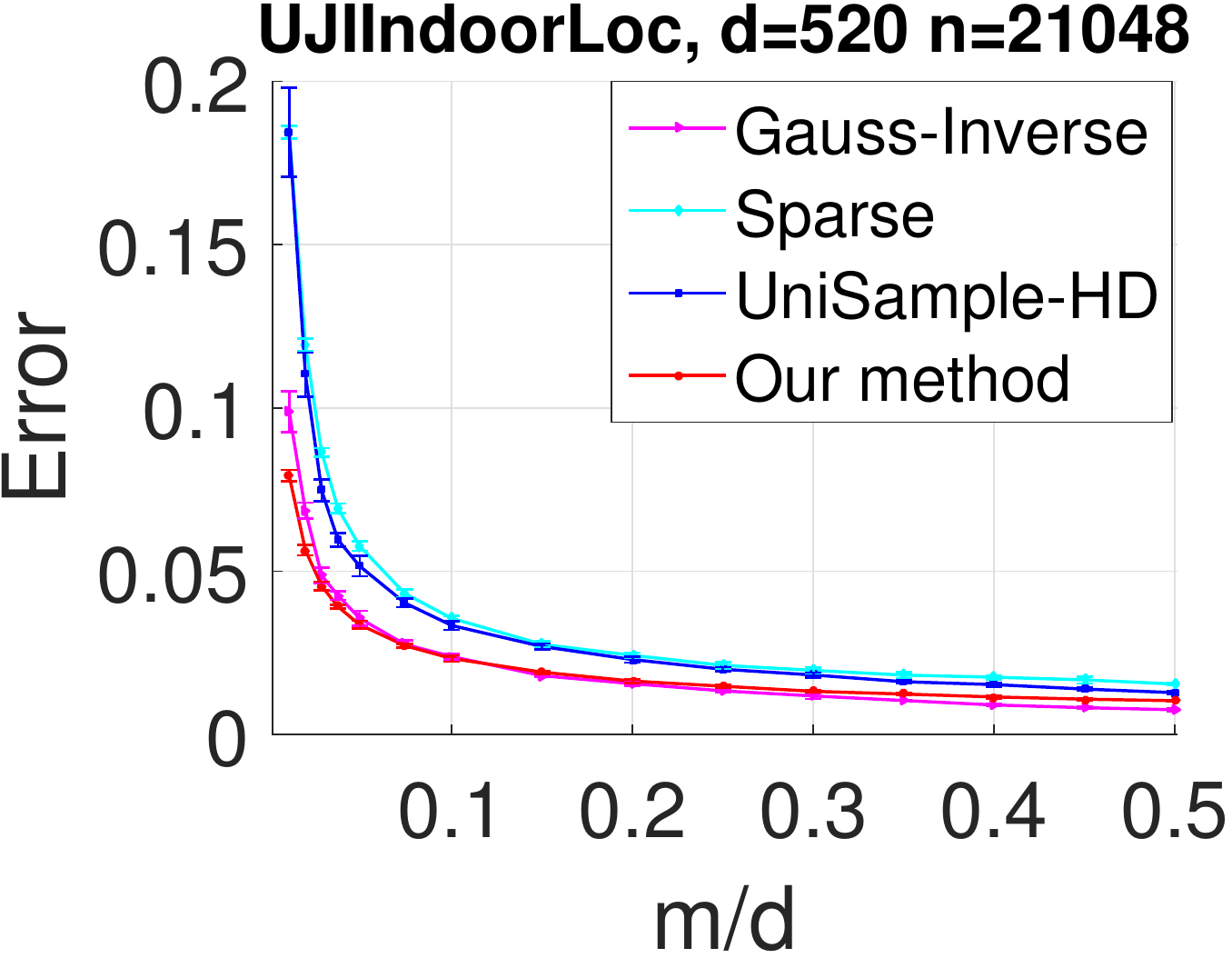} 
}
\subfigure{
\includegraphics[width=.22\textwidth]{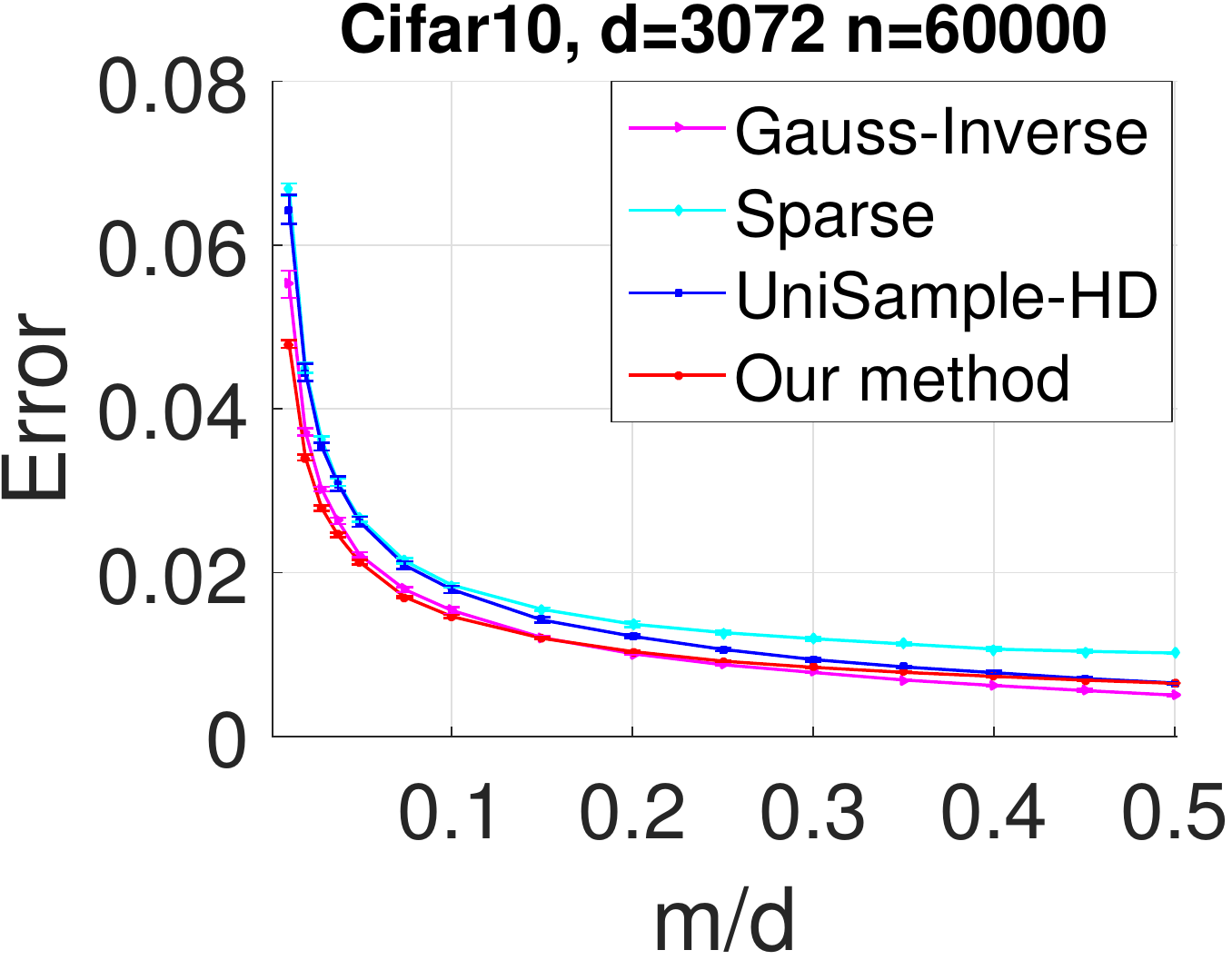}
}
\subfigure{
\includegraphics[width=.22\textwidth]{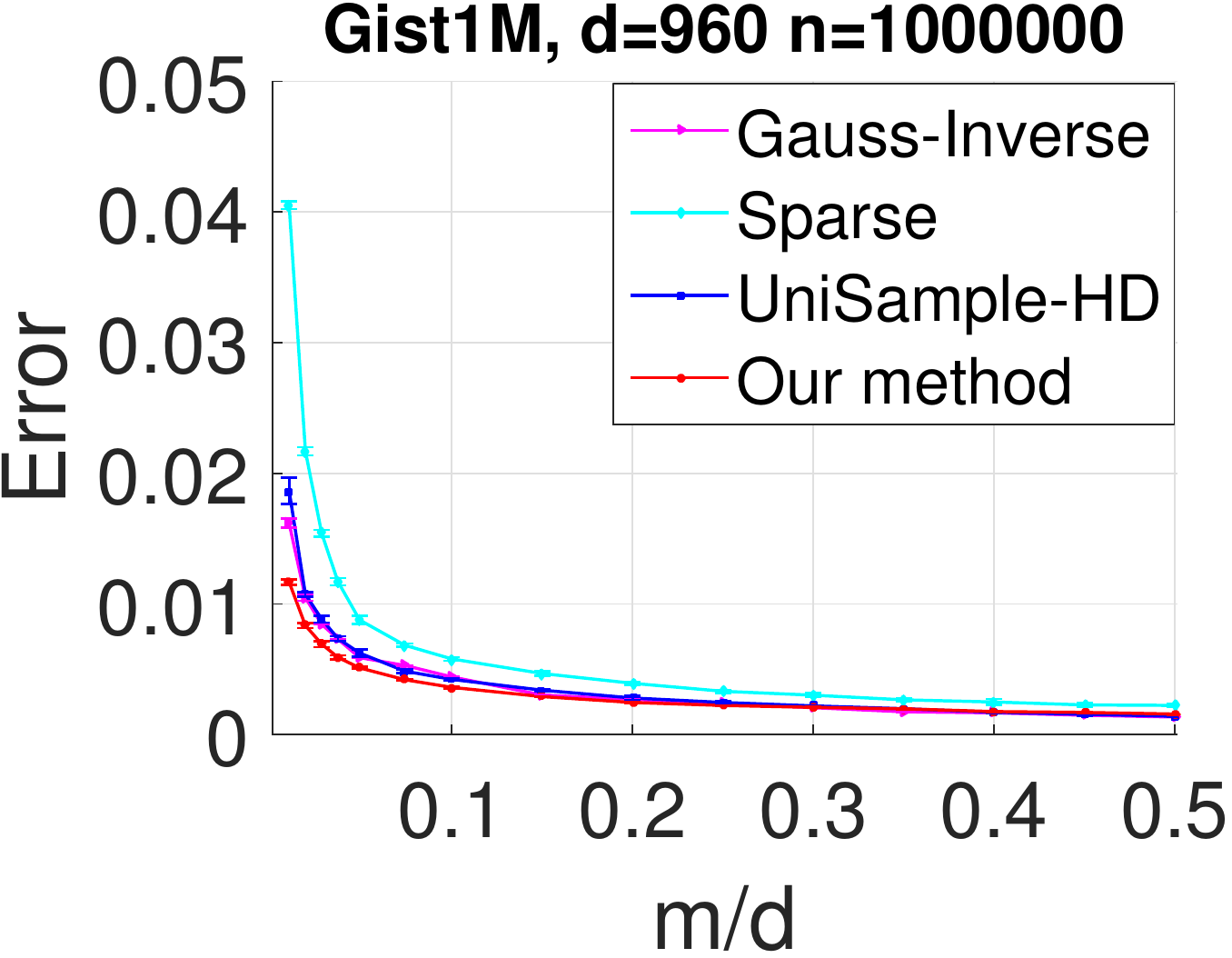}
}
\subfigure{
\includegraphics[width=.22\textwidth]{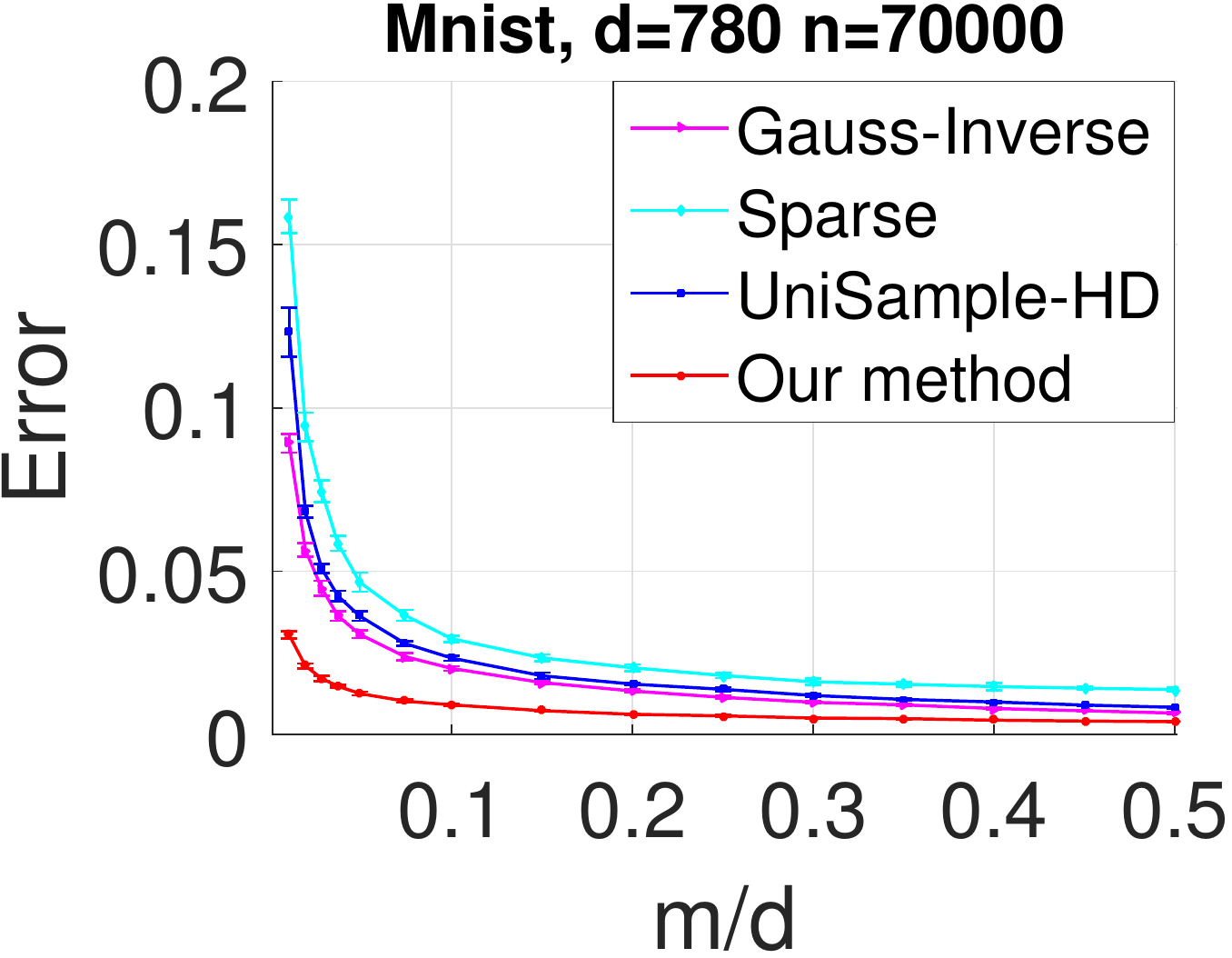}
}
\subfigure{
\includegraphics[width=.22\textwidth]{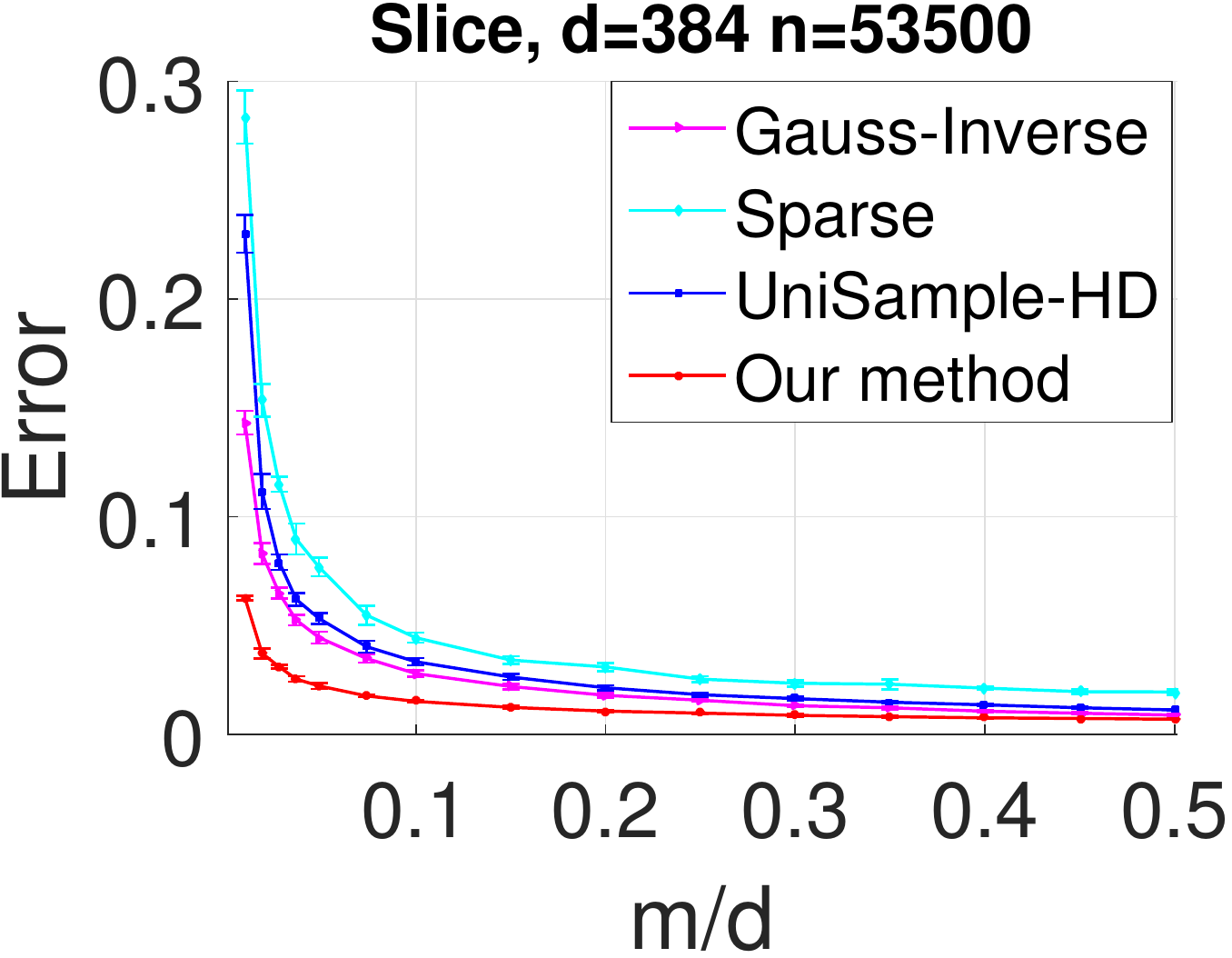} 
}
\caption{Accuracy comparisons of covariance matrix estimation on real-world datasets.}
\label{fig:accuracyrel}
\end{figure} 
 
\subsection{Covariance Estimation on Real-world Datasets}

Figure~\ref{fig:accuracyrel} reports the estimation errors on eight publicly available real-world datasets and shows that the errors decrease dramatically with the increase of $\mathrm{cf}$.  Our DACE consistently exhibits superior accuracy with the least deviation in all cases.

\begin{figure}[htbp]
\centering
\subfigure[]{\label{fig:error_all}
\includegraphics[width=.22\textwidth]{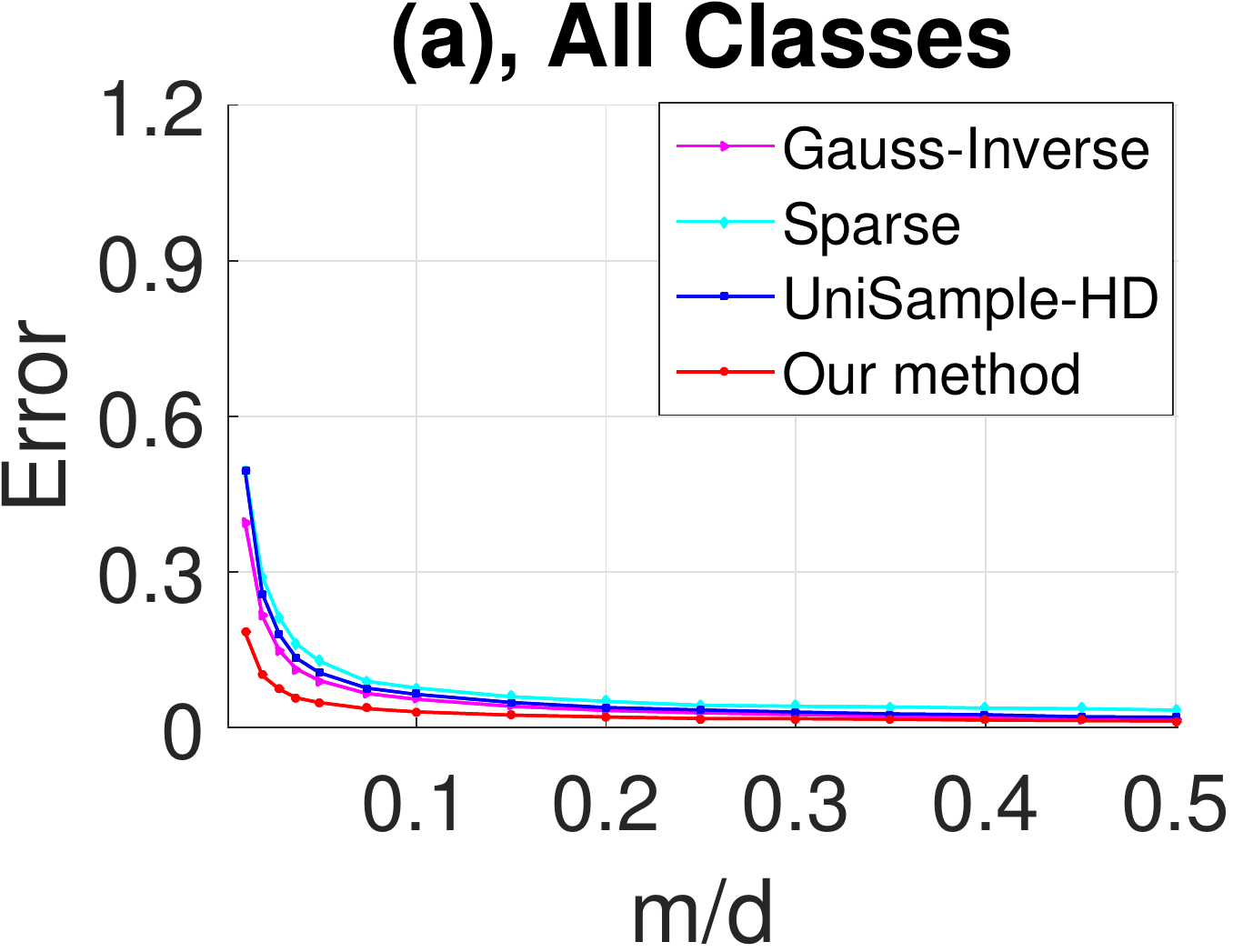}
}
\subfigure[]{\label{fig:error_4}
\includegraphics[width=.22\textwidth]{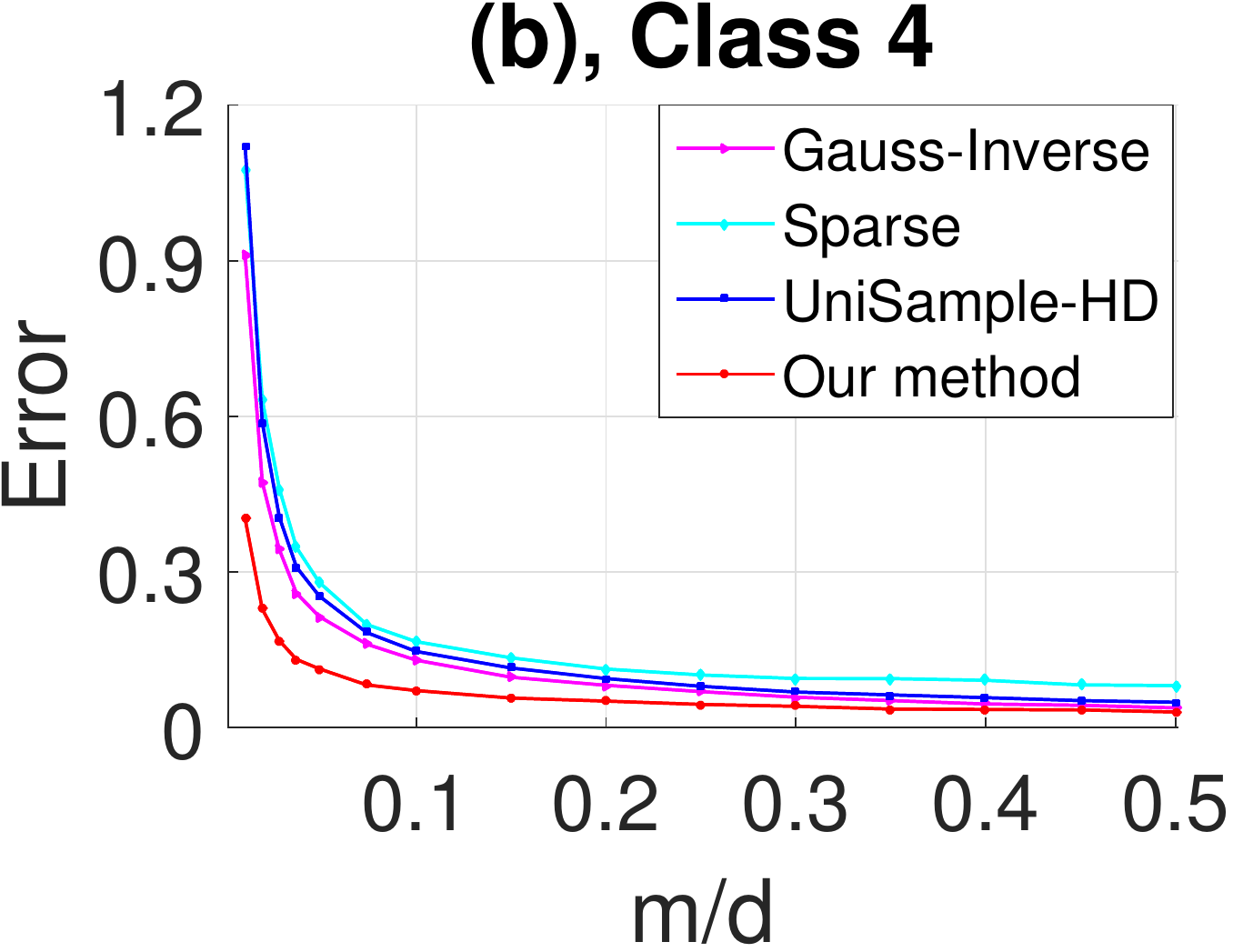} 
}
\subfigure[]{\label{fig:error_8}
\includegraphics[width=.22\textwidth]{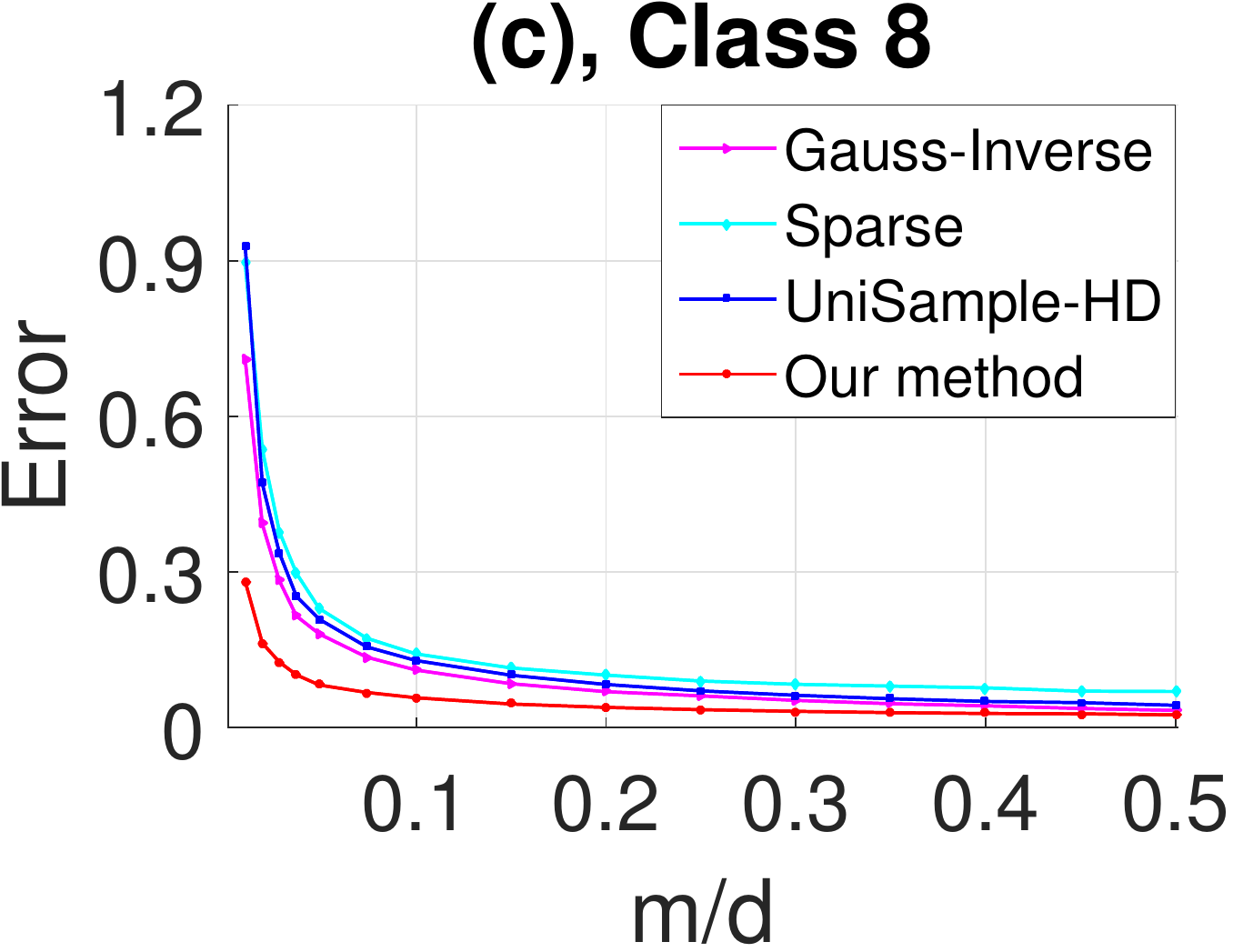}
}
\subfigure[]{\label{fig:error_10}
\includegraphics[width=.22\textwidth]{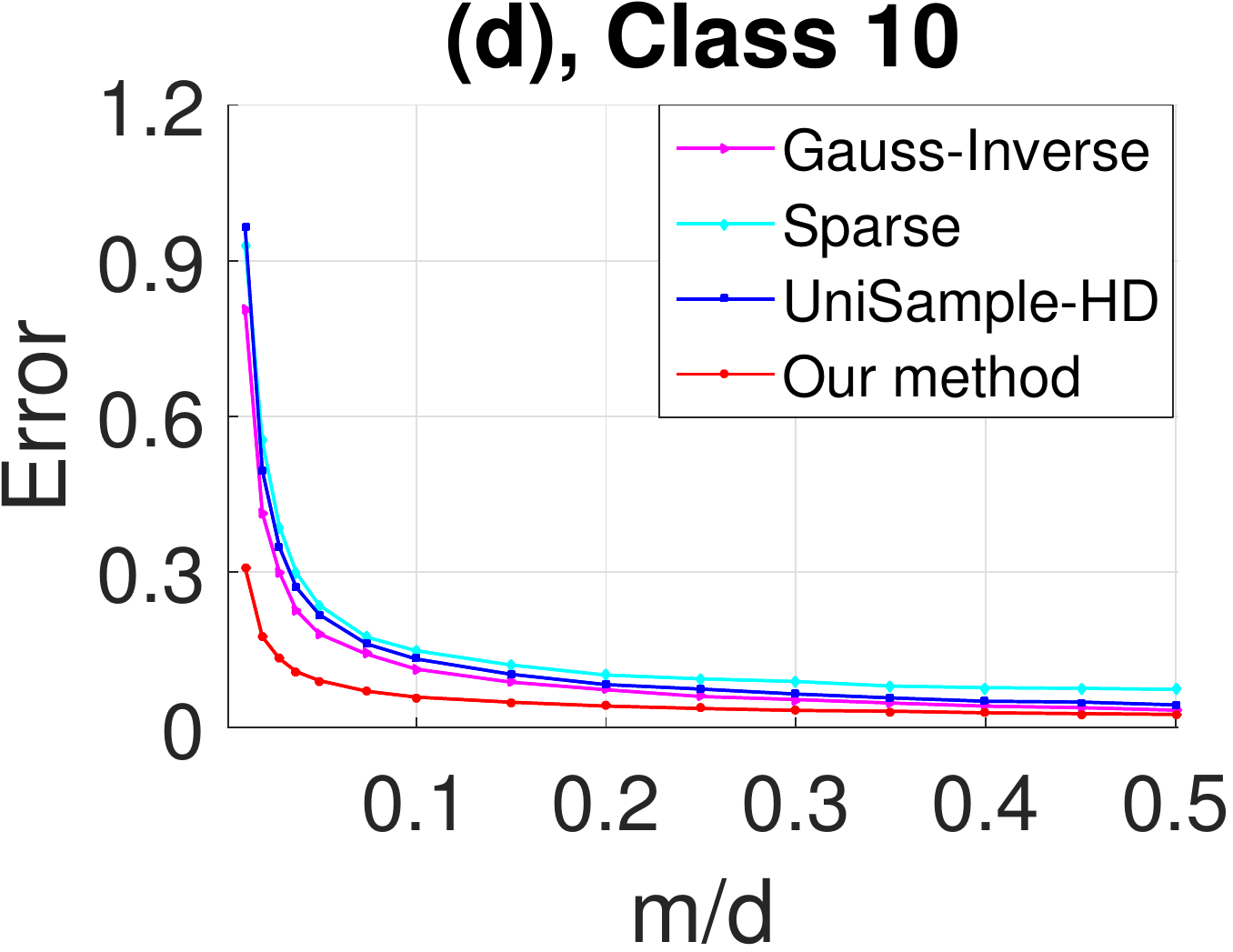} 
}
\subfigure[]{\label{fig:acc_all}
\includegraphics[width=.22\textwidth]{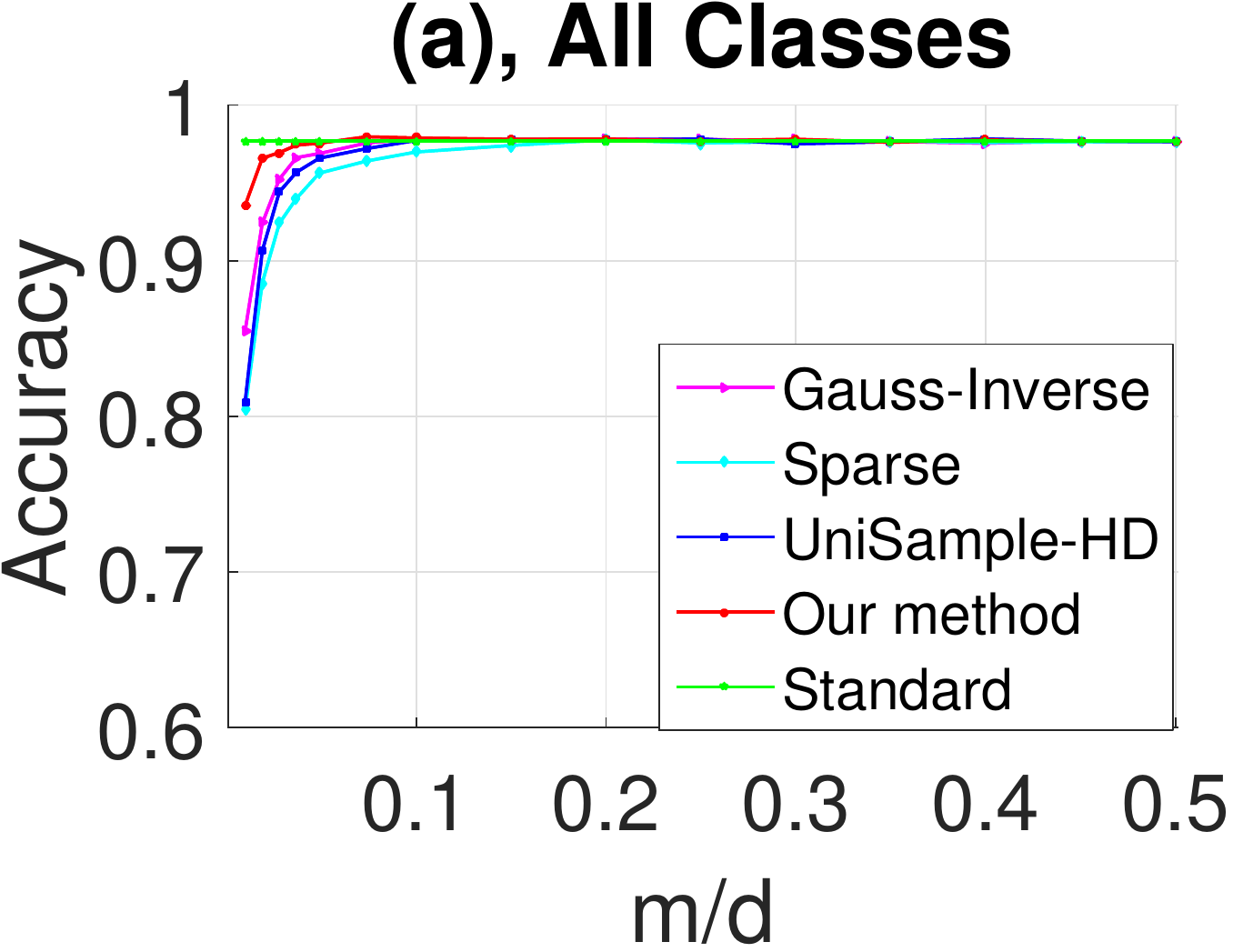}
}
\subfigure[]{\label{fig:acc_4}
\includegraphics[width=.22\textwidth]{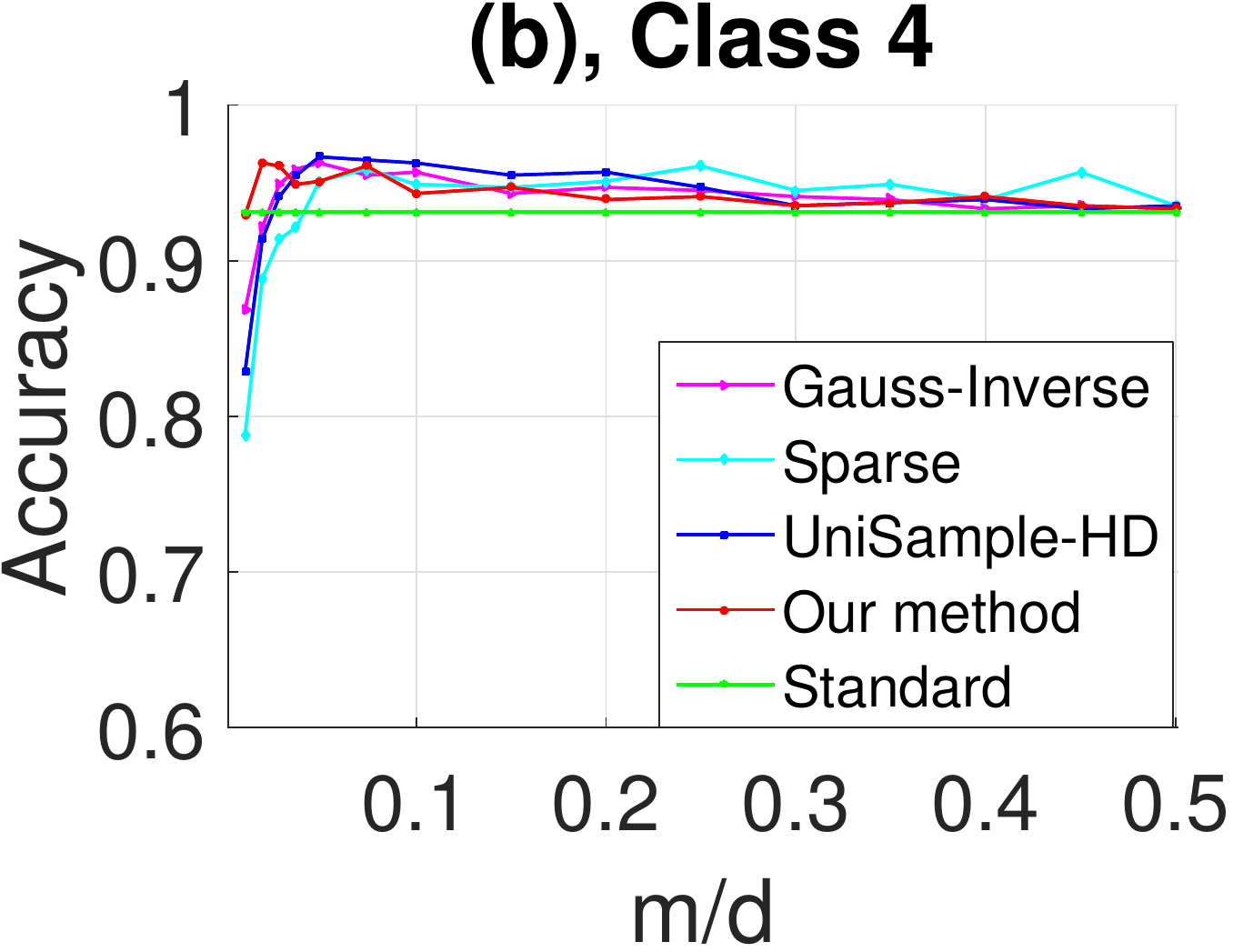} 
}
\subfigure[]{\label{fig:acc_8}
\includegraphics[width=.22\textwidth]{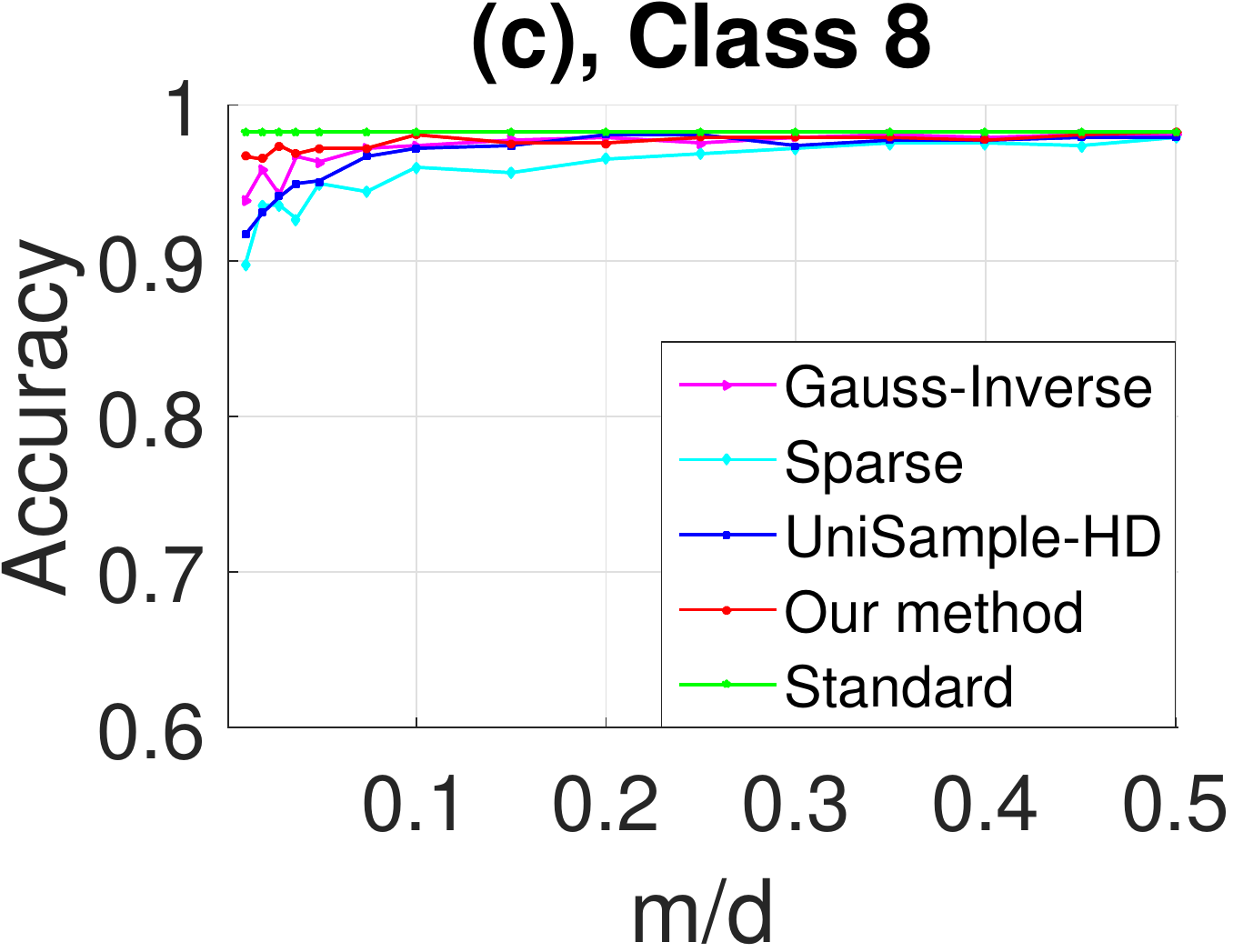}
}
\subfigure[]{\label{fig:acc_10}
\includegraphics[width=.22\textwidth]{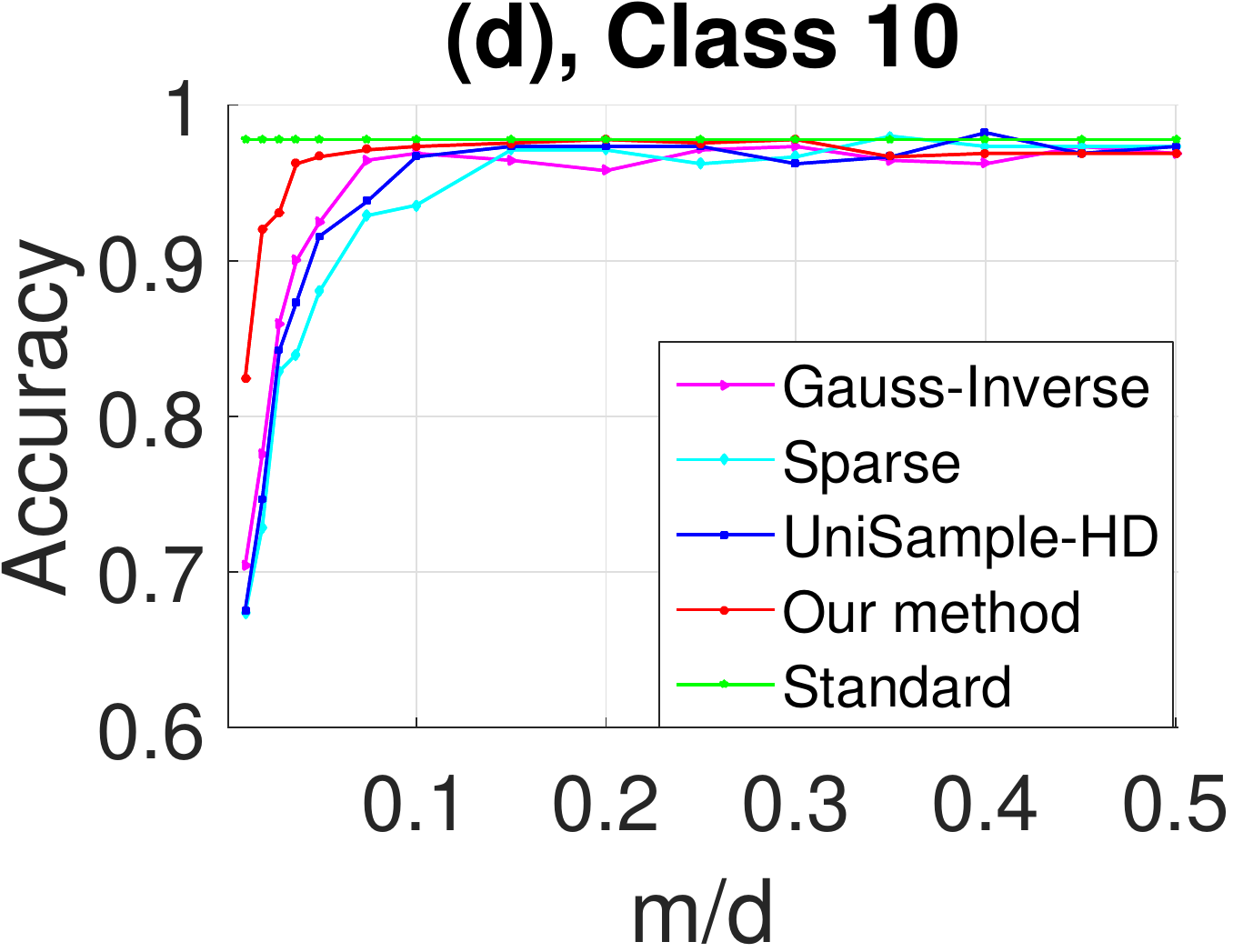} 
}
\caption{{(a) shows the covariance estimation error over all data and  (b)-(d) show the estimation error over the data of three different classes. (e) shows the classification accuracy averaged over all test data and (f)-(h) show the classification accuracy over the test data of three different classes.}}
\label{fig:mnistcla}
\end{figure}

\subsection{Evaluation on Multiclass Classification}


To guarantee that the classification performance is purely determined by the class covariance matrices rather than the mean vectors, we generate a new dataset, namely MNIST-ZM, which centers the MNIST dataset in each class.  The dataset consists of $10$ classes of data with around $7,000$ data points in each class.  $100$ data points from each class are randomly picked for test while the remaining are applied to calculate $\{\C_t\}_{t=1}^{10}$ and $\{\C_{e,t}\}_{t=1}^{10}$.  The parameter $k$ is set to $30$ because \textit{Standard} can obtain good classification results.  The ratio $m/d$ is varied from $0.01$ to $0.5$.

{The left 4 plots in Figure~\ref{fig:mnistcla} present the results of class covariance matrix estimation.  We can observe that the estimation error of each individual class covariance matrix $\C_{e,t}$ is around twice (below $\sqrt{T}=\sqrt{10}$) to that of $\C_{e}$.  Accordingly, the right 4 plots in Figure~\ref{fig:mnistcla} show the compared classification results using the estimated covariance matrices derived from different methods.  We observe that the classification accuracy of our DACE is comparable with \textit{Standard} without performing data compression.  Moreover, our DACE slightly outperforms \textit{Standard} at some $m/d$ value in Fig.~\ref{fig:acc_4}, which depicts that our DACE can select informative features to achieve better generalization. Finally, our DACE also outperforms the other three methods learned from compressed data in terms of both estimation precision and classification accuracy.
}

\section{Conclusion}\label{sec:con}
We present a data-aware weighted sampling method for tackling covariance matrix recovery and multiclass classification problems.  We theoretically prove that our proposed DACE is an unbiased covariance matrix estimator and can employ less data than other representative algorithms to attain the same performance.  The empirical results on both synthetic and real-world datasets support our theory and demonstrate the superior performance over other state-of-the-art methods. 



\appendices
\section{Preliminaries}
The following two lemmas and the recited theorems set the foundation of proving our main theoretical results. 

\begin{lemma} \label{lem:exp}
Given a vector $\x\in \R^d$, sample $m$ ($m<d$) entries from $\x$ with replacement by running Algorithm 1.  Let $\{p_k\}_{k=1}^d$ and   $\S\in\R^{d\times m}$ define as in Algorithm 1.  $\{\e_k\in\R^d\}_{k=1}^d$ denote the standard basis vectors.  Then, we have 
\begin{align}
&\E\left[\S\S^T\x\x^T\S\S^T\right]=\sum_{k=1}^d\frac{x_k^2}{mp_k}\e_k\e_k^T+\frac{m-1}{m}\x\x^T \label{eq:lemma1.1}\\
&\E\left[\diag(\S\S^T\x\x^T\S\S^T)\right]\!\!=\!\!\sum_{k=1}^d(\frac{1}{mp_k}+\frac{m-1}{m})x_k^2\e_k\e_k^T \label{eq:lemma1.2}\\
&\E\left[(\diag(\S\S^T\x\x^T\S\S^T))^2\right] =\sum_{k=1}^d\left[\frac{1}{m^3p_k^3}+\frac{7(m-1)}{m^3p_k^2}\right. \notag \\
&\left.\!\!+\!\!\frac{6(m^2-3m+2)}{m^3p_k}\!\!+\!\!\frac{m^3-6m^2+11m-6}{m^3}\right]x_k^4\e_k\e_k^T\label{eq:lemma1.3}\\
&\E\left[\S\S^T\x\x^T\S\S^T\diag(\S\S^T\x\x^T\S\S^T)\right]\notag\\
=&(\E\left[\diag(\S\S^T\x\x^T\S\S^T)\S\S^T\x\x^T\S\S^T\right])^T\notag\\
=&
\sum_{k=1}^d\left[\frac{1}{m^3p_k^3}+\frac{6(m-1)}{m^3p_k^2}+\frac{3(m^2-3m+2)}{m^3p_k}\right]x_k^4\e_k\e_k^T\notag \\ &\notag +\frac{m-1}{m^3}\x\x^T\diag(\{\frac{x_k^2}{p_k^2}\})+\frac{3(m^2-3m+2)}{m^3}\x\x^T\notag\\ &\cdot\left[\diag(\{\frac{x_k^2}{p_k}\})+\frac{m-3}{3}\diag(\{x_k^2\})\right] \label{eq:lemma1.4}\\
&\E\left[(\S\S^T\x\x^T\S\S^T)^2\right]\!\!=\!\!\sum_{k=1}^d\left[\frac{4(m-1)}{m^3p_k^2}+\frac{1}{m^3p_k^3}\right]x_k^4\e_k\e_k^T\notag\\
&\!\!+\!\!\sum_{k=1}^d\left[\frac{\|\x\|_2^2(m^2-3m+2)}{m^3}+\frac{m-1}{m^3}\sum_{k=1}^d\frac{x_k^2}{p_k}\right]\frac{x_k^2}{p_k}\e_k\e_k^T\notag \\
&\!\!+\!\!\left[\frac{\|\x\|_2^2(m^3\!-\!6m^2\!+\!11m\!-\!6)}{m^3}\!\!+\!\!\frac{m^2\!-\!3m\!+\!2}{m^3}\sum_{k=1}^d\frac{x_k^2}{p_k}\right]\x\x^T\notag \\
&\!\!+\x\x^T\left[\frac{2(m^2-3m+2)}{m^3}\diag(\{\frac{x_k^2}{p_k}\})+\frac{m-1}{m^3}\diag(\{\frac{x_k^2}{p_k^2}\})\right]\notag\\
&\!\!+\!\!\left[\frac{2(m^2-3m+2)}{m^3}\diag(\{\frac{x_k^2}{p_k}\})+\frac{m-1}{m^3}\diag(\{\frac{x_k^2}{p_k^2}\})\right]\x\x^T\label{eq:lemma1.5},
\end{align}
where $\diag(\{x_k^2\})$ denotes a square diagonal matrix with $\{x_k^2\}_{k=1}^d$ on its diagonal and likewise for other notations.
\end{lemma}

\begin{lemma} \label{lem:error}
Following the same notations defined in Lemma~\ref{lem:exp}, with probability at least $1-\sum_{k=1}^d\eta_k$, we have
\begin{eqnarray}
&\|\S\S^T\x\x\S\S^T\|_2\leq\sum_{k\in \Gamma}f^2(x_k,\eta_k,m),
\end{eqnarray}
where $\Gamma$ is a set containing at most $m$ different elements of $[d]$ with its cardinality $|\Gamma|\leq m$ and $f(x_k,\eta_k,m)=|x_k|+\log(\frac{2}{\eta_k})\left[\frac{|x_k|}{3mp_k}+|x_k|\sqrt{\frac{1}{9m^2p_k^2}+\frac{2}{\log(2/\eta_k)}(\frac{1}{mp_k}-\frac{1}{m})}\right]$.
\end{lemma} 


\begin{theorem}[{\cite[p.~76]{tropp2015introduction}}]\label{the:ber}
Let $\{\A_i\}_{i=1}^L\in \R^{d\times n}$ be independent random matrices with  $\E\left[ \A_i\right]=\mathrm{0}$ and $\|\A_i\|_2\leq R$. Define the variance $\sigma^2=\max\{\|\sum_{i=1}^L\E\left[\A_i\A_i^T\right]\|_2, \|\sum_{i=1}^L\E\left[\A_i^T\A_i\right]\|_2\}$. Then,  $\Pr(\|\sum_{i=1}^L\A_i\|_2\geq\epsilon)\leq (d+n)\exp(\frac{-\epsilon^2/2}{\sigma^2+R\epsilon /3})$ for all $\epsilon\geq 0$.
\end{theorem}

\begin{theorem}[{\cite[p.~396]{golub1996matrix}}] \label{the:eigen}
If $\A\in\R^{d\times d}$ and $\A+\Ee\in\R^{d\times d}$ are  symmetric matrices, then
\begin{align}
\lambda_k(\A)+\lambda_d(\Ee)\leq\lambda_k(\A+\Ee)\leq\lambda_k(\A)+\lambda_1(\Ee)
\end{align}
for $k\in [d]$, where $\lambda_k(\A+\Ee)$ and $\lambda_k(\A)$ designate the $k$-th largest eigenvalues.
\end{theorem}

\section{Theoretical Proofs}
\subsection{Proof of Lemma~\ref{lem:error}}
\begin{proof}
According to the notion defined in Lemma~\ref{lem:exp}, we have  
\begin{eqnarray}
&\|\S\S^T\x\x^T\S\S^T\|_2\overset{(a)}{=}\|\S\S^T\x\|_2^2=\|\sum_{j=1}^m\s_{t_j}\s_{t_j}^T\x\|_2^2\notag\\
=&\|\sum_{j=1}^m\frac{1}{mp_{t_j}}x_{t_j}\e_{t_j}\|_2^2=\|\sum_{j=1}^m\sum_{k=1}^d\frac{\delta_{t_jk}}{mp_{k}}x_{k}\e_{k}\|_2^2\notag\\
=&\sum_{k=1}^d(\sum_{j=1}^m\frac{\delta_{t_jk}x_{k}}{mp_{k}})^2\overset{(b)}{=}\sum_{k\in \Gamma}(\sum_{j=1}^m\frac{\delta_{t_jk}x_{k}}{mp_{k}})^2, \label{eq:lemma2step1}
\end{eqnarray}
where $\Gamma=\{\gamma_t\}_{t=1}^{|\Gamma|}$ is a set with the  cardinality $|\Gamma|\leq m$ containing at most $m$ different elements of $[d]$. 

In Eq.~(\ref{eq:lemma2step1}), $(a)$ holds because $\S\S^T\x\x^T\S\S^T$ is a positive semidefinite matrix of rank $1$.  $\delta_{t_jk}$ returns $1$ only when $t_j=k$ and $0$ otherwise.  $ \Pr(\delta_{t_jk}=1)=\Pr(t_j=k)=p_k$.  $(b)$ holds because we perform random sampling with replacement $m$ times on the $d$ entries of $\x\in \R^d$ and consequently at most $m$ different entries from $\x$ are sampled. 

Let $k=\gamma_1, \gamma_1\in\Gamma$, we first bound $|\sum_{j=1}^m\frac{\delta_{t_j{\gamma_1}}x_{\gamma_1}}{mp_{\gamma_1}}|$. Let $a_j=\frac{\delta_{t_j{\gamma_1}}x_{\gamma_1}}{mp_{\gamma_1}}-\frac{x_{\gamma_1}}{m}, j\in [m]$, we can easily check that $\{a_j\}_{j=1}^m$ are independent with $\E\left[a_j\right]=0$, where Theorem~\ref{the:ber} can be applied for our analysis.  Further, we have
\begin{eqnarray}
&\max_{j\in [m]}|a_j|\!\!=\!\!\max\{\frac{|x_{\gamma_1}|}{m}(\frac{1}{p_{\gamma_1}}\!\!-\!\!1),\frac{|x_{\gamma_1}|}{m}\}\!\!\leq\!\! \frac{|x_{\gamma_1}|}{mp_{\gamma_1}}(\!\!\!\!:=\!\!\!\!R),\\
&\rm{and~~}\sum_{j=1}^m\E\left[a_j^2\right]=\frac{x_{\gamma_1}^2}{mp_{\gamma_1}}-\frac{x_{\gamma_1}^2}{m}(\!\!:=\!\!\sigma^2).
\end{eqnarray}
Thus, by applying Theorem~\ref{the:ber}, we obtain  $\Pr(|\sum_{j=1}^ma_j|\!\!\geq\!\!\epsilon)\leq \eta_{\gamma_1}$, where 
$\eta_{\gamma_1}=2\exp(\frac{-\epsilon^2/2}{x_{\gamma_1}^2/(mp_{\gamma_1})\!-\!x_{\gamma_1}^2/m\!+\!|x_{\gamma_1}|\epsilon/(3mp_{\gamma_1})})$.
\!\!\!\! 
Then, with probability at least $1-\eta_{\gamma_1}$, we have $|\sum_{j=1}^ma_j|\leq\epsilon$, i.e., $|\sum_{j=1}^m\frac{\delta_{t_j{\gamma_1}}x_{\gamma_1}}{mp_1}|\leq |x_{\gamma_1}|+\epsilon$.  We then replace $\epsilon$ and obtain the function,  $f(x_{\gamma_1},\eta_{\gamma_1},m)$, defined in Lemma~\ref{rem:2}. 

Similarly, we bound $|\sum_{j=1}^m\frac{\delta_{t_jk}x_k}{mp_k}|$ for  $k\in [d]$.  The lemma holds by using the union bound over cases for all $k \in [d]$.
\end{proof}

\subsection{Proof of Theorem~\ref{the:unbiased}}\label{app:proof_unbiased}
\begin{proof}  
First, we have 
\begin{eqnarray}
&\E[\widehat{\C}_1]=\frac{m}{nm-n}\E\sum_{i=1}^{n}\S_i\S_i^T\x_i\x_i^T\S_i\S_i^T \notag\\
&\mathop{=}^{\rm{by~}(\ref{eq:lemma1.1})}\frac{m}{nm-n}\sum_{i=1}^{n}\left[ \sum_{k=1}^d\frac{x_{ki}^2}{mp_{ki}}\e_k\e_k^T+\frac{m-1}{m}\x_i\x_i^T\right] \notag \\
&=\frac{1}{nm-n}\sum_{i=1}^{n}\sum_{k=1}^d\frac{x_{ki}^2}{p_{ki}}\e_k\e_k^T+\frac{1}{n}\X\X^T. \label{eq:proofbiased}
\end{eqnarray}
Second, by Eq.~(\ref{eq:lemma1.2}) in Lemma~\ref{lem:exp}, we have 
\begin{eqnarray}
\E[\widehat{\C}_2]&=\frac{m}{nm-n}\sum_{i=1}^{n}\E\left[\diag(\S_i\S_i^T\x_i\x_i^T\S_i\S_i^T)\right]\diag(\b_i) \notag\\
&=\frac{1}{nm-n}\sum_{i=1}^{n}\sum_{k=1}^d\frac{x_{ki}^2}{p_{ki}}\e_k\e_k^T. \label{eq:proofbiased1}
\end{eqnarray}
Hence, by Eq.~(\ref{eq:proofbiased}) and Eq.~(\ref{eq:proofbiased1}), we  immediately conclude that $\C_{e}=\widehat{\C}_1-\widehat{\C}_2$ is unbiased for $\C$.
\end{proof}

\subsection{Proof of Theorem~\ref{the:error}}\label{app:proof_error}
\begin{proof}
For simplicity, we define $\A_i=\A_{i_1}-\A_{i_2}-\A_{i_3}$, where $\A_{i_1}=\frac{m\S_i\S_i^T\x_i\x_i^T\S_i\S_i^T}{nm-n}$, $\A_{i_2}=\frac{m\diag(\S_i\S_i^T\x_i\x_i^T\S_i\S_i^T)\diag(\b_i)}{nm-n}$, $\A_{i_3}=\frac{\x_i\x_i^T}{n}$. Then, $\C_e-\C=\sum_{i=1}^n\A_i$. 

Obviously, $\{\A_i\}_{i=1}^n$ are independent zero-mean random matrices.  Hence, Theorem~\ref{the:ber} can be directly applied.  To bound $\|\C_e-\C\|_2$, we then calculate the corresponding parameters $R$ and $\sigma^2$ that characterize the range and variance of $\A_i$, respectively. 

We first derive $R$, i.e., the bound of $\|\A_i\|_2$ for $i\in [n]$.  By expanding $\|\A_i\|_2$, we get
\begin{eqnarray}
& \|\A_i\|_2=\|\A_{i_1}-\A_{i_2}-\A_{i_3}\|_2
\leq \|\A_{i_1}-\A_{i_2}\|_2+\|\A_{i_3}\|_2 \notag \\
& {\leq} \|\A_{i_1}\|_2+\|\A_{i_3}\|_2. \label{eq:hey}
\end{eqnarray}
The last inequality in Eq.~(\ref{eq:hey}) results from 
\begin{eqnarray}
&\quad\|\A_{i_1}-\A_{i_2}\|_2=\max_{k\in [d]}|\lambda_k(\A_{i_1}-\A_{i_2})|\notag \\\notag
&\overset{(a)}{\leq}\max\{|\lambda_{d}(\A_{i_1})-\lambda_{1}(\A_{i_2})|, |\lambda_1(\A_{i_1})-\lambda_d(\A_{i_2})|\}\\\notag
&\overset{(b)}{=}\max\{\lambda_{1}(\A_{i_2}), |\lambda_1(\A_{i_1})-\lambda_d(\A_{i_2})|\}\\\notag
&\overset{(c)}{=}\max\{\lambda_{1}(\A_{i_2}), \lambda_1(\A_{i_1})-\lambda_d(\A_{i_2}) \}\\\notag
&\overset{(d)}{\leq}\lambda_1(\A_{i_1})\overset{(e)}{=}\|\A_{i_1}\|_2,
\end{eqnarray}
where $\lambda_k(\cdot)$ is the $k$-th largest eigenvalue. The inequality (a) holds because $\lambda_k(\A_{i_1})-\lambda_1(\A_{i_2})\leq\lambda_k(\A_{i_1}-\A_{i_2})\leq\lambda_k(\A_{i_1})-\lambda_d(\A_{i_2})$ for any $k\in [d]$, which is attained by applying Theorem~\ref{the:eigen} with the fact that  $\lambda_d(-\A_{i_2})=-\lambda_1(\A_{i_2})$ and $\lambda_1(-\A_{i_2})=-\lambda_d(\A_{i_2})$ for $\A_{i_2}\in\R^{d\times d}$.  The equality (b) holds because $\lambda_{k\geq 2}(\A_{i_1})=0$ from the fact that $\A_{i_1}$ is a positive semidefinite matrix of rank $1$ and $\lambda_{k\in [d]}(\A_{i_2})\geq 0$ since $\A_{i_2}$ is positive semidefinite.  The equality (c) follows the fact that $\lambda_1(\A_{i_1})=\Tr(\A_{i_1})\geq\Tr(\A_{i_2})=\sum_{k=1}^d\lambda_k(\A_{i_2})\geq\lambda_d(\A_{i_2})\geq 0$, where the first equality holds because $\lambda_{k\geq 2}(\A_{i_1})=0$, the first inequality results from the fact that the diagonal matrix $\A_{i_2}$ is constructed by the diagonal elements of $\A_{i_1}$ multiplied by positive scalars not bigger than $1$, and the second inequality is the consequence of $\lambda_{k\in [d]}(\A_{i_2})\geq 0$.  The equality (d) results from that $\lambda_{k\in [d]}(\A_{i_2})\geq 0$.  The equality (e) holds due to the fact that $\A_{i_1}$ is positive semidefinite.

Now, we only need to bound $\|\A_{i_1}\|_2$ and $\|\A_{i_3}\|_2$. We have 
\begin{eqnarray}
\|\A_{i_3}\|_2=\|\frac{\x_i\x_i^T}{n}\|_2=\frac{\|\x_i\|_2^2}{n}. 
\end{eqnarray}
Applying Lemma~\ref{lem:error} gets with probability at least $1-\sum_{k=1}^d\eta_{ki}$,
\begin{eqnarray}
\|\A_{i_1}\|_2\leq\frac{m}{nm-n}\sum_{k\in \Gamma_i}f^2(x_{ki},\eta_{ki},m),
\end{eqnarray} 
where  $\Gamma_i=\{\gamma_{ti}\}_{t=1}^{|\Gamma_i|}$ is a set occupying at most $m$ different elements of $[d]$ with its cardinality $|\Gamma_i|\leq m$ and $f(x_{ki},\eta_{ki},m)=|x_{ki}|+\log(\frac{2}{\eta_{ki}})\left[\frac{|x_{ki}|}{3mp_{ki}}+|x_{ki}|\sqrt{\frac{1}{9m^2p_{ki}^2}+\frac{2}{\log(2/\eta_{ki})}(\frac{1}{mp_{ki}}-\frac{1}{m})}\right]$. 

We can derive similar bounds for all $\{\x_i\}_{i=1}^n$.  Then, by applying the union bound, with probability at least $1-\sum_{i=1}^n\sum_{k=1}^d\eta_{ki}$, we have
\begin{eqnarray}
R=\max_{i\in [n]}\left[\frac{m}{nm-n}\sum_{k\in\Gamma_i}f^2(x_{ki},\eta_{ki},m)+\frac{\|\x_i\|_2^2}{n}\right].\label{eq:th2R}
\end{eqnarray}
Applying the inequality $(\sum_{t=1}^n a_t)^2\leq n\sum_{t=1}^n a_t^2$, we have 
\begin{align}
&f^2(x_{ki},\eta_{ki},m)\leq  3x_{ki}^2+3\log^2(\frac{2}{\eta_{ki}})\frac{x_{ki}^2}{9m^2p_{ki}^2}\notag\\&\quad+3\log^2(\frac{2}{\eta_{ki}})\frac{x_{ki}^2}{9m^2p_{ki}^2}+6\log(\frac{2}{\eta_{ki}})(\frac{x_{ki}^2}{mp_{ki}}-\frac{x_{ki}^2}{m})\notag \\
&\leq 3x_{ki}^2+\log^2(\frac{2}{\eta_{ki}})\frac{2x_{ki}^2}{3m^2p_{ki}^2}+\log(\frac{2}{\eta_{ki}})\frac{6x_{ki}^2}{mp_{ki}}.\label{eq:th2R1}
\end{align}
Before continuing characterizing $R$ in Eq.~(\ref{eq:th2R}), we set the sampling probabilities as $p_{ki}=\alpha\frac{|x_{ki}|}{\|\x_i\|_1}+(1-\alpha)\frac{x_{ki}^2}{\|\x_i\|_2^2}$. It is easy to check that $\sum_{k=1}^dp_{ki}=1$. For $0<\alpha<1$,  we also have $p_{ki}\geq\alpha\frac{|x_{ki}|}{\|\x_i\|_1}$, then plugging it in the second and the third term of Eq.~(\ref{eq:th2R1}) respectively, we get
\begin{align}
&\!\!\!\! f^2(x_{ki},\eta_{ki},m)\leq U_{ki}\notag\\ &\!\!\!\!U_{ki}\!\!\!:=\!\!\!3x_{ki}^2\!+\!\log^2(\frac{2}{\eta_{ki}})\frac{2\|\x_i\|_1^2}{3m^2\alpha^2}\!+\!\log(\frac{2}{\eta_{ki}})\frac{6|x_{ki}|\|\x_i\|_1}{m\alpha}.\label{eq:th2R2}
\end{align}
Equipped with Eq.~(\ref{eq:th2R}) and setting $\eta_{ki}=\frac{\eta}{nd}$ for all $i\in [n]$ and $k\in[d]$, we bound $R$ with probability at least $1-\sum_{i=1}^n\sum_{k=1}^d\eta_{ki}= 1-\eta$ by
\begin{eqnarray}
R& \leq \max_{i\in [n]}\left[\frac{m}{nm-n}\sum_{k\in\Gamma_i}U_{ki}+\frac{\|\x_i\|_2^2}{n}\right]\notag \\
&\leq \max_{i\in [n]}\left[\frac{2}{n}\Big(3\|\x_i\|_2^2+\log^2(\frac{2nd}{\eta})\frac{2\|\x_i\|_1^2}{3m\alpha^2}\right.\notag\\
&\quad\left.+\log(\frac{2nd}{\eta})\frac{6\|\x_i\|_1^2}{m\alpha}\Big)+\frac{\|\x_i\|_2^2}{n}\right]\notag \\
&\leq \max_{i\in [n]}\left[\frac{7\|\x_i\|_2^2}{n}+\log^2(\frac{2nd}{\eta})\frac{14\|\x_i\|_1^2}{nm\alpha^2}\right],\label{eq:th2R3}
\end{eqnarray}
where the second inequality follows from that $\frac{m}{m-1}\leq 2$ for $m\geq 2$ and $|\Gamma_i|\leq m$ and the last inequality results from that $\alpha\leq 1$ and $\log(\frac{2nd}{\eta})\geq 1$ for $n\geq 1$, $d\geq 2$, and $\eta\leq 1$. 

We then derive $\sigma^2$ by only bounding for $\|\sum_{i=1}^n\E\left[\A_i\A_i\right]\|_2$ since $\A_i$ is symmetric.  By expanding $\E\left[\A_i\A_i\right]$, we obtain
\begin{eqnarray}
&&\!\!\!\!\mathbf{0}\preceq \E\left[\A_i\A_i\right]=\E\left[\A_{i_1}\A_{i_1}+\A_{i_2}\A_{i_2}+\A_{i_3}\A_{i_3}-\A_{i_1}\A_{i_2} \right.\notag \\ 
&&\!\!\!\!\left.
-\A_{i_2}\A_{i_1}-\A_{i_1}\A_{i_3}-\A_{i_3}\A_{i_1}+\A_{i_2}\A_{i_3}+\A_{i_3}\A_{i_2} \right] \notag
\end{eqnarray}
In the following, we bound the expectation of each term. Specifically, invoking Lemma~\ref{lem:exp}, we have
\begin{eqnarray}
&&\!\!\!\!n^2\E\left[\A_i\A_i\right]=\sum_{i=1}^{11}\textcircled{\scriptsize i}-\sum_{i=12}^{22}\textcircled{\scriptsize i}, \quad \rm{where}\label{eq:longexpression}\\
\notag &&\!\!\!\!\textcircled{\scriptsize 1}\!\!:=\!\!\sum_{k=1}^d\left[\frac{4}{m(m-1)p_{ki}^2}+\frac{1}{(m-1)^2mp_{ki}^3}\right]x_{ki}^4\e_k\e_k^T\\\notag
&&\!\!\!\!\textcircled{\scriptsize 2}\!\!:=\!\!\sum_{k=1}^d\left[\frac{\|\x_i\|_2^2(m-2)}{m(m-1)}+\frac{1}{m(m-1)}\sum_{k=1}^d\frac{x_{ki}^2}{p_{ki}}\right]\frac{x_{ki}^2}{p_{ki}}\e_k\e_k^T\\\notag&&\!\!\!\!
\textcircled{\scriptsize 3}\!\!:=\!\!\left[\frac{\|\x_i\|_2^2(m^2-5m+6)}{m(m-1)}+\frac{m-2}{m(m-1)}\sum_{k=1}^d\frac{x_{ki}^2}{p_{ki}}\right]\x_i\x_i^T\\\notag&&\!\!\!\!
\textcircled{\scriptsize 4}\!\!:=\!\!\frac{2(m-2)}{m(m-1)}\x_i\x_i^T\diag(\{\frac{x_{ki}^2}{p_{ki}}\}), \textcircled{\scriptsize 5}\!\!:=\!\!\frac{1}{m(m-1)}\x_i\x_i^T\diag(\{\frac{x_{ki}^2}{p_{ki}^2}\})\\\notag&&\!\!\!\!
\textcircled{\scriptsize 6}\!\!:=\!\!\frac{2(m-2)}{m(m-1)}\diag(\{\frac{x_{ki}^2}{p_{ki}}\})\x_i\x_i^T, \textcircled{\scriptsize 7}\!\!:=\!\!\frac{1}{m(m-1)}\diag(\{\frac{x_{ki}^2}{p_{ki}^2}\})\x_i\x_i^T\\\notag&&\!\!\!\!
\textcircled{\scriptsize 8}\!\!:=\!\!\diag(\b_i)\diag(\b_i)
\sum_{k=1}^d\left[\frac{1}{m(m-1)^2p_{ki}^3}\!+\!\frac{7}{m(m-1)p_{ki}^2}\right.\\\notag&&\!\!\!\!
\quad\quad\left.+\!\frac{6(m-2)}{m(m-1)p_{ki}}\!+\!\frac{(m-2)(m-3)}{m(m-1)}\right]x_{ki}^4\e_k\e_k^T\\\notag&&\!\!\!\!
\textcircled{\scriptsize 9}\!\!:=\!\!\|\x_i\|_2^2\x_i\x_i^T, \textcircled{\scriptsize 10}\!\!\!\!:=\!\!\!\!\sum_{k=1}^d(\frac{1}{(m-1)p_{ki}}+1)x_{ki}^2\e_k\e_k^T\diag(\b_i)\x_i\x_i^T \\\notag&&\!\!\!\!
\textcircled{\scriptsize 11}\!\!:=\!\!\x_i\x_i^T\sum_{k=1}^d(\frac{1}{(m-1)p_{ki}}+1)x_{ki}^2\e_k\e_k^T\diag(\b_i)\\\notag&&\!\!\!\!
\textcircled{\scriptsize 12}\!\!:=\!\!2\sum_{k=1}^d\left[\frac{1}{m(m-1)^2p_{ki}^3}+\frac{6}{m(m-1)p_{ki}^2}\right.\\\notag &&\!\!\!\!\left.\qquad+\frac{3(m-2)}{m(m-1)p_{ki}}\right]x_{ki}^4\e_k\e_k^T\diag(\b_i)\\\notag&&\!\!\!\!
\textcircled{\scriptsize 13}\!\!:=\!\!\frac{3(m-2)}{m(m-1)}\x_i\x_i^T\diag(\{\frac{x_{ki}^2}{p_{ki}}\})\diag(\b_i)\\\notag&&\!\!\!\!
\textcircled{\scriptsize 14}\!\!:=\!\!\frac{(m-2)(m-3)}{m(m-1)}\x_i\x_i^T\diag(\{x_{ki}^2\})\diag(\b_i)\\\notag&&\!\!\!\!
\textcircled{\scriptsize 15}\!\!:=\!\!\frac{3(m-2)}{m(m-1)}\diag(\b_i)\diag(\{\frac{x_{ki}^2}{p_{ki}}\})\x_i\x_i^T\\\notag&&\!\!\!\!
\textcircled{\scriptsize 16}\!\!:=\!\!\frac{(m-2)(m-3)}{m(m-1)}\diag(\b_i)\diag(\{x_{ki}^2\})\x_i\x_i^T\\\notag&&\!\!\!\!
\textcircled{\scriptsize 17}\!\!:=\!\!\sum_{k=1}^d\frac{x_{ki}^2}{(m-1)p_{ki}}\e_k\e_k^T\x_i\x_i^T,\quad \textcircled{\scriptsize 18}\!\!:=\!\!\|\x_i\|_2^2\x_i\x_i^T\\\notag &&\!\!\!\!
\textcircled{\scriptsize 19}\!\!:=\!\!\sum_{k=1}^d\frac{x_{ki}^2}{(m-1)p_{ki}}\x_i\x_i^T\e_k\e_k^T, \quad \textcircled{\scriptsize 20}\!\!:=\!\!\|\x_i\|_2^2\x_i\x_i^T\\\notag&&\!\!\!\!
\textcircled{\scriptsize 21}\!\!:=\!\!\frac{1}{m(m-1)}\x_i\x_i^T\diag(\{\frac{x_{ki}^2}{p_{ki}^2}\})\diag(\b_i)\\\notag&&\!\!\!\!
\textcircled{\scriptsize 22}\!\!:=\!\!\frac{1}{m(m-1)}\diag(\b_i)\diag(\{\frac{x_{ki}^2}{p_{ki}^2}\})\x_i\x_i^T.
\end{eqnarray}
In Eq.~(\ref{eq:longexpression}), for $m\geq 2$, we have
\begin{align}\notag
&\textcircled{\scriptsize 10}-\textcircled{\scriptsize 17} = \mathbf{0},\quad 
\textcircled{\scriptsize 11}-\textcircled{\scriptsize 19}=\mathbf{0}\\
& \textcircled{\scriptsize 4}-\textcircled{\scriptsize 13}+\textcircled{\scriptsize 5}-\textcircled{\scriptsize 14}-\textcircled{\scriptsize 21}\notag\\
\!=\!&\frac{\x_i\x_i^T}{m(m\!-\!1)}\diag(\{\frac{((m-1)/p_{ki})x_{ki}^2}{1+(m-1)p_{ki}}\!\!+\!\!\frac{(m\!-\!2)(m\!\!+\!\!1\!-\!1/p_{ki})x_{ki}^2}{1+(m-1)p_{ki}}\})\notag \\\notag
&\textcircled{\scriptsize 6}-\textcircled{\scriptsize 15}+\textcircled{\scriptsize 7}-\textcircled{\scriptsize 16}-\textcircled{\scriptsize 22}\\\notag
=&\diag(\{\frac{((m\!-\!1)/p_{ki})x_{ki}^2}{1\!+\!(m\!-\!1)p_{ki}}\!+\!\frac{(m\!-\!2)(m\!+\!1\!-\!1/p_{ki})x_{ki}^2}{1+(m-1)p_{ki}}\})\frac{\x_i\x_i^T}{m(m\!-\!1)}\notag \\\notag
&\textcircled{\scriptsize 3}+\textcircled{\scriptsize 9}-\textcircled{\scriptsize 18}-\textcircled{\scriptsize 20}\\=&\left[\frac{(6\!-\!4m)\|\x_i\|_2^2}{m^2-m}\!+\!\frac{m-2}{m^2-m}\sum_{k=1}^d\frac{x_{ki}^2}{p_{ki}}\right]\x_i\x_i^T
\preceq \sum_{k=1}^d\frac{x_{ki}^2}{mp_{ki}}\x_i\x_i^T;\notag \\
&\textcircled{\scriptsize 8}-\textcircled{\scriptsize 12}\preceq \mathbf{0}; \notag \\
&\textcircled{\scriptsize 1} \preceq \sum_{k=1}^d\left[\frac{8x_{ki}^4}{m^2p_{ki}^2}+\frac{4x_{ki}^4}{m^3p_{ki}^3}\right]\e_{k}\e_{k}^T; \notag \\
&\textcircled{\scriptsize 2} \preceq \sum_{k=1}^d\left[\frac{\|\x_i\|_2^2x_{ki}^2}{mp_{ki}}+\frac{2x_{ki}^2}{m^2p_{ki}}\sum_{k=1}^d\frac{x_{ki}^2}{p_{ki}} \right]\e_k\e_k^T .\label{eq:longsimplify}
\end{align}
Then, by applying Eq.~(\ref{eq:longexpression}) and Eq.~(\ref{eq:longsimplify}), we obtain
\begin{align}\notag
&\mathbf{0} \preceq \E\left[\A_i\A_i\right]\\\notag
& \!\preceq\! \frac{1}{n^2}\sum_{k=1}^d\left[\frac{8x_{ki}^4}{m^2p_{ki}^2}\!\!+\!\!\frac{4x_{ki}^4}{m^3p_{ki}^3}\!\!+\!\!\frac{\|\x_i\|_2^2x_{ki}^2}{mp_{ki}}\!\!+\!\!\frac{2x_{ki}^2}{m^2p_{ki}}\sum_{k=1}^d\frac{x_{ki}^2}{p_{ki}}\right]\e_{k}\e_{k}^T \notag\\ 
&+\frac{\x_i\x_i^T}{n^2m(m\!-\!1)}\diag(\{\frac{{\scriptscriptstyle{m-1\over p_{ki}}}x_{ki}^2}{1\!+\!(m\!-\!1)p_{ki}}\!+\!\frac{(m\!-\!2)(m\!+\!1\!-\!{\scriptscriptstyle {1\over p_{ki}}})x_{ki}^2}{1+(m-1)p_{ki}}\})\notag\\
&+\diag(\{\frac{((m-1)/p_{ki})x_{ki}^2}{1+(m-1)p_{ki}}+\frac{(m-2)(m+1-1/p_{ki})x_{ki}^2}{1+(m-1)p_{ki}}\})\notag \\
&\quad \cdot\frac{\x_i\x_i^T}{n^2m(m-1)}+\frac{1}{n^2m}\sum_{k=1}^d\frac{x_{ki}^2}{p_{ki}}\x_i\x_i^T.
\label{eq:th2t2}
\end{align}
With Eq.~(\ref{eq:th2t2}) in hand, we can formulate $\sigma^2$ as
\begin{align}\notag
&\sigma^2 =\|\sum_{i=1}^n\E\left[\A_i\A_i\right]\|_2\\
&\leq \sum_{i=1}^n\max_{k\in [d]}\frac{1}{n^2}\!\!\left[\frac{8x_{ki}^4}{m^2p_{ki}^2}\!\!+\!\!\frac{4x_{ki}^4}{m^3p_{ki}^3}\!\!+\!\!\frac{\|\x_i\|_2^2x_{ki}^2}{mp_{ki}}\!\!+\!\!\frac{2x_{ki}^2}{m^2p_{ki}}\sum_{k=1}^d\frac{x_{ki}^2}{p_{ki}} \right]\notag
\\\notag
&+\sum_{i=1}^n\max_{k\in [d]}\frac{1}{n^2}\left[\frac{2\|\x_i\|_2^2}{m(m-1)}(\frac{((m-1)/p_{ki})x_{ki}^2}{1+(m-1)p_{ki}}\right.\\&\left.+\frac{(m-2)(m+1+1/p_{ki})x_{ki}^2}{1+(m-1)p_{ki}}) \right]\!+\!\frac{1}{n^2m}\|\sum_{i=1}^n\sum_{k=1}^d\frac{x_{ki}^2}{p_{ki}}\x_i\x_i^T\|_2\notag \\
&\leq \sum_{i=1}^n\max_{k\in [d]}\frac{1}{n^2}\!\left[\frac{8x_{ki}^4}{m^2p_{ki}^2}\!\!+\!\!\frac{4x_{ki}^4}{m^3p_{ki}^3}\!\!+\!\!\frac{\|\x_i\|_2^2x_{ki}^2}{mp_{ki}}\!\!+\!\!\frac{2x_{ki}^2}{m^2p_{ki}}\sum_{k=1}^d\frac{x_{ki}^2}{p_{ki}} \right]\notag
\\
&\!\!+\!\!\sum_{i=1}^n\max_{k\in [d]}\frac{1}{n^2}\!\!\left[ \frac{8\|\x_i\|_2^2 x_{ki}^2}{mp_{ki}} \right]\!\!+\!\!\frac{1}{n^2m}\!\|\!\sum_{i=1}^n\!\sum_{k=1}^d\frac{x_{ki}^2}{p_{ki}}\x_i\x_i^T\|_2. \label{eq:th2t3}
\end{align}
As $p_{ki}=\alpha\frac{|x_{ki}|}{\|\x_i\|_1}+(1-\alpha)\frac{x_{ki}^2}{\|\x_i\|_2^2}$ with $0<\alpha <1$, by plugging $p_{ki}\geq\alpha\frac{|x_{ki}|}{\|\x_i\|_1}$ and $p_{ki}\geq(1-\alpha)\frac{x_{ki}^2}{\|\x_i\|_2^2}$ into  Eq.~(\ref{eq:th2t3}), we have 
\begin{align}\notag 
&\sigma^2 \leq \sum_{i=1}^n\max_{k\in [d]} \frac{1}{n^2}\left[\frac{8\|\x_i\|_2^4}{m^2(1-\alpha)^2}+\frac{4\|\x_i\|_1^2\|\x_i\|_2^2}{m^3\alpha^2(1-\alpha)}+\frac{\|\x_i\|_2^4}{m(1-\alpha)}\right.\\\notag 
&\left. +\frac{2\|\x_i\|_2^2}{m^2(1-\alpha)}\sum_{k=1}^d\frac{|x_{ki}|\|\x_i\|_1}{\alpha} \right]+\sum_{i=1}^n\max_{k\in [d]}\frac{1}{n^2}\left[\frac{8\|\x_i\|_2^4}{m(1-\alpha)} \right]\\\notag & \quad+\frac{1}{n^2m}\|\sum_{i=1}^n\sum_{k=1}^d\frac{|x_{ki}|\|\x_i\|_1}{\alpha}\x_i\x_i^T\|_2 \notag \\
&=\sum_{i=1}^n\left[\frac{8\|\x_i\|_2^4}{n^2m^2(1-\alpha)^2}+\frac{4\|\x_i\|_1^2\|\x_i\|_2^2}{n^2m^3\alpha^2(1-\alpha)}+\frac{9\|\x_i\|_2^4}{n^2m(1-\alpha)}\right. \notag \\
&\quad\left.+\frac{2\|\x_i\|_2^2\|\x_i\|_1^2}{n^2m^2\alpha(1-\alpha)} \right]+ \|\sum_{i=1}^n\frac{\|\x_i\|_1^2\x_i\x_i^2}{n^2m\alpha}\|_2.
\label{eq:th2t4}
\end{align}

By invoking Theorem~\ref{the:ber}, we obtain that for $\epsilon\geq 0$,
\begin{align}
\Pr(\|\C_e-\C\|_2\geq\epsilon)\leq 2d\exp(\frac{-\epsilon^2/2}{\sigma^2+R\epsilon /3}) (:=\delta)\label{eq:theorem}, 
\end{align}
and the following quadratic equation in $\epsilon$: 
\begin{align}
\frac{\epsilon^2}{2\log(2d/\delta)}-\frac{R\epsilon}{3}-\sigma^2=0\label{eq:theorem1}.
\end{align}
Solving Eq.~(\ref{eq:theorem1}), we get the positive root: 
\begin{align}
\epsilon &=\log(\frac{2d}{\delta})\left[\frac{R}{3}+\sqrt{(\frac{R}{3})^2+\frac{2\sigma^2}{\log(2d/\delta)}}\right] \notag\\
&\leq \log(\frac{2d}{\delta})\frac{2R}{3}+\sqrt{2\sigma^2\log(\frac{2d}{\delta})}\label{eq:theorem2}.
\end{align}
Thus, $\|\C_e-\C\|_2\leq\log(\frac{2d}{\delta})\frac{2R}{3}+\sqrt{2\sigma^2\log(\frac{2d}{\delta})}$ holds with probability at least $1-\eta-\delta$ and we complete the proof.
\end{proof}
\subsection{Proof of Corollary~\ref{cor:spec}}\label{app:proof_spec}
\begin{proof} 
By setting $\|\x_i\|_2\leq\tau$ for all $i\in [n]$,  $\varphi:=\frac{\|\x_i\|_1}{\|\x_i\|_2}$, where $1\leq \varphi\leq\sqrt{d}$ and $m<d$ into Theorem~\ref{the:error}, we obtain
\begin{align}\notag
&\|\C_e-\C\|_2\\\leq& \widetilde{O}\Big(\frac{\tau^2}{n}\!+\!\frac{\tau^2\varphi^2}{nm}\!+\!\sqrt{\frac{\tau^4}{nm^2}\!+\!\frac{\tau^4\varphi^2}{nm^3}\!+\!\frac{\tau^4}{nm}\!+\!\frac{\tau^4\varphi^2}{nm^2}\!+\!\frac{\|\C\|_2\tau^2\varphi^2}{nm}}\Big) \notag \\ 
\leq&\widetilde{O}\Big(\frac{\tau^2}{n}\!+\!\frac{\tau^2\varphi^2}{nm}\!+\!\frac{\tau^2\varphi}{m}\sqrt{\frac{1}{n}}\!+\!\tau^2\sqrt{\frac{1}{nm}}\!+\!\tau\varphi\sqrt{\frac{\|\C\|_2}{nm}}\Big),
\label{eq:corollary1proof1}
\end{align}
where the first inequality invokes $\sum_{i=n}^n\|\x_i\|_2^4\leq n\tau^4$ and $\C=\sum_{i=1}^n\frac{\x_i\x_i^T}{n}$ is the original covariance matrix. 

We can adopt $\sum_{i=1}^n\|\x_i\|_2^4\leq nd\tau^2\|\C\|_2$, which holds because $\sum_{i=1}^n\|\x_i\|_2^4\leq \tau^2\sum_{i=1}^n\|\x_i\|_2^2$ and $\sum_{i=1}^n\|\x_i\|_2^2=n\Tr(\C)\leq nd\|\C\|_2$, and derive
\begin{align}\notag
&\|\C_e-\C\|_2\\\leq& \widetilde{O}\Big(\frac{\tau^2}{n}\!\!+\!\!\frac{\tau^2\varphi^2}{nm}\!+\!\tau\sqrt{\!\|\C\|_2\!}\sqrt{\!\frac{d}{nm^2}\!+\!\frac{d\varphi^2}{nm^3}\!+\!\frac{d}{nm}\!+\!\frac{d\varphi^2}{nm^2}\!+\!\frac{\varphi^2}{nm}}\!\Big) \notag \\ 
\leq&\widetilde{O}\Big(\frac{\tau^2}{n}\!\!+\!\!\frac{\tau^2\varphi^2}{nm}\!\!+\!\!\frac{\tau\varphi}{m}\sqrt{\!\frac{d\|\C\|_2}{n}\!}\!\!+\!\!\tau\sqrt{\!\frac{d\|\C\|_2}{nm}\!}\!\!+\!\!\tau\varphi\sqrt{\!\frac{\|\C\|_2}{nm}}\!\Big)
\label{eq:corollary1proof2}  
\end{align}
Finally, assigning the smaller one of Eq.~(\ref{eq:corollary1proof1}) and Eq.~(\ref{eq:corollary1proof2}) to $\|\C_e-\C\|_2$ completes the proof.
\end{proof} 

\subsection{Proof of Corollaries~\ref{cor:gau} and~\ref{cor:subspace}}\label{app:proof_corollaries}
\begin{proof}
The proof follows~[\cite{azizyan2015extreme},~Corollaries 4-6], where the key component $\|\C_e-\C_p\|_2$ is upper bounded by $\|\C_e-\frac{1}{n}\sum_{i=1}^n\x_i\x_i^T\|_2+\|\frac{1}{n}\sum_{i=1}^n\x_i\x_i^T-\C_p\|_2$. Then, via Theorem~\ref{the:error} and the Gaussian tail bounds in~[\cite{azizyan2015extreme},~Proposition 14], we can show that with probability at least $1-\zeta$ for $d\geq 2$, 
\begin{align}
&\max_{i\in [n]}\|\x_i\|_2 \leq\sqrt{2\Tr(\C_p)\log(nd/\zeta)}\notag \\
&\|\frac{1}{n}\sum_{i=1}^n\x_i\x_i^T-\C_p\|_2\leq O\big( \|\C_p\|_2\sqrt{\log(2/\zeta)/n}\big). \label{eq:concentrations}
\end{align}
Applying them and Corollary~\ref{cor:spec} along with the fact that $\|\x_i\|_1\leq\sqrt{d}\|\x_i\|_2$ and $\Tr(\C_p)\leq d\|\C_p\|_2$, we establish 
\begin{align}
&\|\C_e-\C_p\|_2\leq\|\C_e-\frac{1}{n}\sum_{i=1}^n\x_i\x_i^T\|_2+\|\frac{1}{n}\sum_{i=1}^n\x_i\x_i^T-\C_p\|_2 \notag \\
&\leq\widetilde{O}\Big(\frac{\tau^2}{n}\!\!+\!\!\frac{\tau^2\varphi^2}{nm}\!\!+\!\!\frac{\tau^2\varphi}{m}\sqrt{\!\frac{1}{n}}\!\!+\!\!\tau^2\sqrt{\!\frac{1}{nm}}\!+\!\tau\varphi\sqrt{\!\frac{\|\frac{1}{n}\!\sum_{i=1}^n\x_i\x_i^T\|_2}{nm}}\Big) \notag \\\notag&
\qquad + \widetilde{O}\Big(\|\C_p\|_2\sqrt{\frac{1}{n}}\Big)\\
&\leq \widetilde{O}\Big(\frac{d^2\|\C_p\|_2}{nm}+\frac{d\|\C_p\|_2}{m}\sqrt{\frac{d}{n}}\Big) \label{eq:corollary2proof1}
\end{align}
with probability at least $1-\eta-\delta-\zeta$, where  we invoke Eq.~(\ref{eq:concentrations}) to get  $\|\frac{1}{n}\sum_{i=1}^n\x_i\x_i^T\|_2\leq \|\frac{1}{n}\sum_{i=1}^n\x_i\x_i^T-\C_p\|_2+\|\C_p\|_2\leq\widetilde{O}(\|\C_p\|_2)$.

Let rank($\C_p$)$\leq r$, we have the result for the low-rank case 
\begin{align}
\|[\C_e]_r-\C_p\|_2 & \leq\!\! \|[\C_e]_r-\C_e\|_2 + \|\C_e-\C_p\|_2  \notag \\
&\leq\!\! \|[\C_p]_r-\C_e\|_2 + \|\C_e-\C_p\|_2  \notag \\
&\leq \!\!\|[\C_p]_r\!-\!\C_p\|_2\! +\! \|\C_p\!-\!\C_e\|_2\! +\! \|\C_e\!-\!\C_p\|_2  \notag \\
& =\!\! 2\|\C_e-\C_p\|_2 \label{eq:corollary2proof21},
\end{align}
where the last equality holds because $\text{rank}(\C_p)\leq r$. Then, armed with $\Tr(\C_p)\leq \text{rank}(\C_p)\|\C_p\|_2 \leq r \|\C_p\|_2$, we have 
\begin{align}
&\quad\|[\C_e]_r-\C_p\|_2\leq O(\|\C_e-\C_p\|_2)\notag\\
&\leq O(\|\C_e-\frac{1}{n}\sum_{i=1}^n\x_i\x_i^T\|_2+\|\frac{1}{n}\sum_{i=1}^n\x_i\x_i^T-\C_p\|_2)\notag \\
&\leq\!\! \widetilde{O}\Big(\frac{rd\|\C_p\|_2}{nm}\!+\!\frac{r\|\C_p\|_2}{m}\sqrt{\frac{d}{n}}\!+\!\|\C_p\|_2\sqrt{\frac{rd}{nm}}\Big)\label{eq:corollary2proof2}
\end{align}
with probability at least $1-\eta-\delta-\zeta$.

Due to the symmetry of $\C_p$ and $\C_e$, following~\cite{azizyan2015extreme}, we can combine Davis-Kahan Theorem~\cite{davis1970rotation},   $\|\widehat{\prod}_k-\prod_k\|_2\leq \frac{1}{\lambda_k-\lambda_{k+1}}\|\C_e-\C_p\|_2$, with the result from Corollary~\ref{cor:gau} and immediately derive the desired bound in Corollary~\ref{cor:subspace}. 

\end{proof}

\bibliographystyle{abbrv}
\bibliography{example_paper}
\end{document}